\documentclass[preprint]{isipta2025}
\usepackage{tikz} 
\usepackage{nicefrac}

\definecolor{definecolour}{RGB}{45,125,162} 

\newcommand*{\DummyArgumentSymbol}{\hbox{\scalebox{1.25}{\(\bullet\)}}}
\newcommand*{\bolleke}{\DummyArgumentSymbol}
\providecommand\given{}

\newcommand\AltCondSymbol[1][]{\nonscript\,#1\Vert\allowbreak\nonscript\,\mathopen{}}
\newcommand\NewInfoSymbol[1][]{\nonscript\,#1\vert\allowbreak\nonscript\,\mathopen{}}

\newcommand*{\DesirSymbol}{D}
\newcommand*{\IndifSymbol}{I}
\newcommand*{\AssessmentSymbol}{A}
\newcommand*{\ModelSymbol}{M}
\newcommand*{\BackgroundModelSymbol}{V}
\newcommand*{\PowerSetSymbol}{\mathscr{P}}
\newcommand*{\OptionsSetSymbol}{U}
\newcommand*{\EventsSetSymbol}{E}
\newcommand*{\GamblesSetSymbol}{\mathscr{G}}
\newcommand*{\ExpansionSymbol}{E}
\newcommand*{\RevisionSymbol}{R}
\newcommand*{\ContractionSymbol}{C}

\DeclareMathOperator{\posi}{posi}
\DeclareMathOperator{\shull}{sh}
\DeclareMathOperator{\linspan}{span}
\DeclareMathOperator{\clsr}{cl}
\DeclareMathOperator{\comp}{co}
\DeclareMathOperator{\range}{rng}

\DeclarePairedDelimiterX\group[1]{(}{)}{\ifblank{#1}{\,\DummyArgumentSymbol\,}{#1}}
\DeclarePairedDelimiterX{\set}[1]{\{}{\}}{\renewcommand\given{\SetSymbol}#1}
\DeclarePairedDelimiter{\structure}{\langle}{\rangle}
\DeclarePairedDelimiterX{\Adelim}[2]{\langle}{\rangle}{\,#1\,;\,#2\,}

\DeclarePairedDelimiterXPP{\compof}[1]{\comp}{(}{)}{}{\ifblank{#1}{\,\DummyArgumentSymbol\,}{#1}}
\DeclarePairedDelimiterXPP{\shullof}[1]{\shull}{(}{)}{}{\ifblank{#1}{\,\DummyArgumentSymbol\,}{#1}}
\DeclarePairedDelimiterXPP{\posiof}[1]{\posi}{(}{)}{}{\ifblank{#1}{\,\DummyArgumentSymbol\,}{#1}}
\DeclarePairedDelimiterXPP{\linspanof}[1]{\linspan}{(}{)}{}{\ifblank{#1}{\,\DummyArgumentSymbol\,}{#1}}
\DeclarePairedDelimiterXPP{\closureof}[2]{\clsr_{#1}}{(}{)}{}{\ifblank{#2}{\,\DummyArgumentSymbol\,}{#2}}
\DeclarePairedDelimiterXPP{\cohmodclosureof}[1]{\cohmodclosure}{(}{)}{}{\ifblank{#1}{\,\DummyArgumentSymbol\,}{#1}}

\DeclarePairedDelimiterXPP{\rangeof}[1]{\range}{(}{)}{}{\ifblank{#1}{\,\DummyArgumentSymbol\,}{#1}}
\DeclarePairedDelimiterXPP{\powersetof}[1]{\PowerSetSymbol}{(}{)}{}{\ifblank{#1}{\,\DummyArgumentSymbol\,}{#1}}
\DeclarePairedDelimiterXPP{\expandof}[1]{\expand}{(}{)}{}{\renewcommand\given{\NewInfoSymbol}#1}
\DeclarePairedDelimiterXPP{\reviseof}[1]{\revise}{(}{)}{}{\renewcommand\given{\NewInfoSymbol}#1}
\DeclarePairedDelimiterXPP{\contractof}[1]{\contract}{(}{)}{}{\renewcommand\given{\NewInfoSymbol}#1}

\DeclarePairedDelimiterXPP{\Msabove}[1]{\Ms}{(}{)}{}{\ifblank{#1}{\,\DummyArgumentSymbol\,}{#1}}
\DeclarePairedDelimiterXPP{\closedMsabove}[1]{\closedMs}{(}{)}{}{\ifblank{#1}{\,\DummyArgumentSymbol\,}{#1}}
\DeclarePairedDelimiterXPP{\difMsabove}[1]{\difMs}{(}{)}{}{\ifblank{#1}{\,\DummyArgumentSymbol\,}{#1}}

\newcommand*{\cls}[1]{\clsr_{#1}}
\newcommand*{\union}{\cup}
\newcommand*{\intersection}{\cap}
\newcommand*{\Intersection}{\bigcap}
\newcommand*{\Mcls}{\cls{\Ms}}

\newcommand*{\Mclsof}[2][]{\closureof[#1]{\Ms}{#2}}
\newcommand*{\Mclsbgof}[3][]{\closureof[#1]{\Msabove{#2}}{#3}}

\newcommand*{\reals}{\mathbb{R}}

\newcommand*{\then}{\Rightarrow}
\newcommand*{\ifandonlyif}{\Leftrightarrow}

\newcommand*{\opt}[1][]{u_{#1}}
\newcommand*{\altopt}[1][]{v_{#1}}

\newcommand*{\eventopt}[1][]{e_{#1}}
\newcommand*{\calledoff}[2][]{\eventopt[#1]\ast#2}
\newcommand*{\uniteventopt}{1_\opts}
\newcommand*{\nulleventopt}{0_\opts}

\newcommand*{\opts}{\mathscr{\OptionsSetSymbol}}
\newcommand*{\eventopts}{\mathscr{\EventsSetSymbol}}

\newcommand*{\posopts}{\opts_{\optgt0}}
\newcommand*{\eventposopts}[1][]{\opts_{\eventoptgt[#1]0}}

\newcommand*{\desirset}[1][]{\DesirSymbol_{#1}}
\newcommand*{\gbl}{f}

\newcommand*{\gambles}{\mathscr{\GamblesSetSymbol}}

\newcommand{\varstate}[1][]{{\mathit{X}_{#1}}}
\newcommand*{\states}[1][]{\mathscr{X}_{#1}}
\newcommand*{\state}[1][]{x_{#1}}
\newcommand*{\event}[1][]{E_{#1}}
\newcommand*{\indof}[1]{\mathbb{I}_{#1}}
\newcommand*{\indevent}[1][]{\indof{\event[#1]}}
\newcommand*{\belmod}[1][]{b_{#1}}
\newcommand*{\belmods}{\mathbf{B}}
\newcommand*{\cohmod}[1][]{c_{#1}}
\newcommand*{\cohmods}{\mathbf{C}}
\newcommand*{\closedmods}{\overline{\cohmods}}
\newcommand*{\belmodtop}{\mathbf{1}_{\belmods}}
\newcommand*{\belmodbottom}{\mathbf{0}_{\belmods}}
\newcommand*{\modleq}{\sqsubseteq}
\newcommand*{\cohmodbottom}{\mathbf{0}_{\cohmods}}
\newcommand*{\cohmodclosure}{\clsr_{\cohmods}}

\newcommand*{\expand}{\mathrm{\ExpansionSymbol}}
\newcommand*{\revise}{\mathrm{\RevisionSymbol}}
\newcommand*{\contract}{\mathrm{\ContractionSymbol}}

\newcommand*{\A}{\AssessmentSymbol}
\newcommand*{\otherA}{B}
\newcommand*{\As}{\mathbf{\AssessmentSymbol}}
\newcommand{\Astop}{\mathbf{1}_{\As}}

\newcommand*{\M}{\ModelSymbol}
\newcommand*{\Ms}{\mathbf{\ModelSymbol}}
\newcommand*{\closedMs}{\smash{\overline{\Ms}}}
\newcommand*{\difMs}{\dif\Ms}

\newcommand*{\bgM}{\BackgroundModelSymbol}
\newcommand*{\zeroM}{{\bgM_o}}
\newcommand*{\eventbgM}{{\bgM_{\eventopt}}}
\newcommand*{\eventM}[1][\eventopt]{\M_{#1}}
\newcommand*{\indexedeventM}[1][]{\M_{\eventopt[#1]}}
\newcommand*{\dif}[1]{#1_{\textrm{\tiny\upshape{DI}}}\vphantom{#1}}

\newcommand*{\des}{\smallblacktriangleright}
\newcommand*{\rej}{\vartriangleleft}
\newcommand*{\acc}{\trianglerighteq}
\newcommand*{\indif}{\equiv}
\newcommand*{\unres}{\smile}

\newcommand*{\posetleq}{\sqsubseteq}

\newcommand*{\meet}{\frown}
\newcommand*{\join}{\smile}

\newcommand*{\optlt}[1][]{\mathrel{\prec_{#1}}}
\newcommand*{\optgt}[1][]{\mathrel{\succ_{#1}}}
\newcommand*{\eventoptgt}[1][]{\optgt[{\eventopt[#1]}]}
\newcommand*{\optleq}[1][]{\mathrel{\preceq_{#1}}}
\newcommand*{\optgeq}[1][]{\mathrel{\succeq_{#1}}}

\newcommand*{\weakgt}{>}
\newcommand*{\weakgeq}{\geq}

\newcommand*{\indifset}[1][]{\IndifSymbol_{#1}}
\newcommand*{\eventindifset}[1][]{\indifset[{\eventopt[#1]}]}
\newcommand*{\altquotient}[2][\indifset]{{#2}\!/{#1}}
\newcommand*{\quotientopts}{\altquotient{\opts}}
\newcommand*{\altcondon}[1]{\AltCondSymbol{#1}}

\newcommand*{\define}[1]{\emph{\textcolor{definecolour}{#1}}\/}
\newcommand*{\nameit}[1]{\totheright{\upshape\textcolor{definecolour}{[#1]}}}
\newcommand*{\totheright}[1]{{\unskip\nobreak\hfil\penalty50\hskip1em\hbox{}\nobreak\hfil#1\parfillskip=0pt\finalhyphendemerits=0\par}}
\newcommand*{\instantiateas}{\rightsquigarrow}

\DeclarePairedDelimiter\ket{\lvert}{\rangle}
\DeclarePairedDelimiterX\braket[2]{\langle}{\rangle}{#1\delimsize\vert#2}
\DeclarePairedDelimiterX\braketwithop[3]{\langle}{\rangle}{#1\delimsize\vert#2\delimsize\vert#3}

\DeclareMathOperator{\spec}{spec}

\newcommand{\hilbertspace}{\mathscr{X}}
\newcommand{\statespace}{\bar{\hilbertspace}}
\newcommand{\subspace}[1][]{\mathscr{W}_{#1}}
\newcommand{\fket}[1][]{\smash{\ket{\phi_{#1}}}}
\newcommand{\uket}{\ket{\Psi}}

\newcommand{\operator}[1]{\hat{#1}}
\newcommand{\measurement}[1]{\operator{#1}}
\newcommand{\measurements}{\mathscr{H}}
\newcommand{\projectoron}[1]{\measurement{P}_{#1}}
\newcommand{\projector}[1][]{\projectoron{\subspace[#1]}}
\newcommand{\identity}{\measurement{I}}
\newcommand{\zero}{\measurement{0}}
\newcommand{\spectrum}[1]{\spec(\measurement{#1})}

\DeclareRobustCommand{\citefirstlastauthor}{\AtNextCite{\DeclareNameAlias{labelname}{given-family}}\citeauthor}

\newif\ifarXiv
\arXivtrue 

\title{Conditioning and AGM-like belief change\\in the Desirability-Indifference framework}

\author[1]{Kathelijne Coussement}
\author[1]{Gert de Cooman}
\author[1]{Keano De Vos}

\affil[1]{Ghent University, Foundations Lab for imprecise probabilities, Ghent, Belgium}

\begin{document}

\maketitle


\begin{abstract}
We show how the AGM framework for belief change (expansion, revision, contraction) can be extended to deal with conditioning in the so-called Desirability-Indifference framework, based on abstract notions of accepting and rejecting options, as well as on abstract notions of events.
This level of abstraction allows us to deal simultaneously with classical and quantum probability theory.
\end{abstract}



\section{The problem and its context}\label{sec::introduction}
In an earlier paper \cite{cooman2003a}, \citefirstlastauthor{cooman2003a} proposed a generalisation of the AGM framework for belief change --- change of belief state --- in propositional logic (and probability theory) \cite{gardenfors1988} to what he called \emph{belief models}, which are meant as a much more general and abstract representation for what may constitute a belief state.

More recently, he and co-authors introduced a theory of accepting and rejecting gambles \cite{quaeghebeur2015:statement}, which provides a quite general, well-reasoned and operationalisable framework for decision-making under uncertainty, and provides a context for dealing with such notions as acceptability, desirability, and indifference of gambles, as well as the relationships between them.

In this paper, which is intended as a first exploration, we combine ideas from both papers to show how the AGM framework for belief change can be extended to deal with conditioning on events in the so-called Desirability-Indifference framework, based on abstract notions of accepting and rejecting options (uncertain rewards), as well as on abstract notions of events.

The paper is structured as follows.
In \cref{sec::accept:reject}, we introduce the general framework of \emph{statement models} for abstract options, which are meant to represent uncertain rewards.
We pay particular attention to a specific type of statement models, which define what we call the \emph{Desirability-Indifference (or DI) framework}.
A generalised notion of conditioning, based on abstract events that can be used to call off options, is briefly discussed in \cref{sec::events}, and \cref{sec::belief:change} has a succinct summary of the extension of the AGM framework for belief change to belief models.
With this scaffolding in place, we show in \cref{sec::conditioning} how statement models fit into the belief model framework, and how our abstract notion of conditioning can be seen as a form of belief change --- belief revision --- in the Desirability-Indifference framework, satisfying relevant generalised versions of the AGM axioms, as well as versions of the so-called Levi and Harper identities.
\Cref{sec::instances} very briefly discusses two specific instances of the general framework, namely classical and quantum probabilistic inference, and in the Conclusion we point to some ways to further develop and refine our theory.

We have collected proofs for all the (arguably) less trivial claims we make in an Appendix.

\section{Statement models}\label{sec::accept:reject}
We consider an agent, called You, who is asked to make accept and/or reject statements about options.
\define{Options} are abstract objects that represent uncertain rewards.
When You accept an option, You agree to receiving an uncertain reward\footnote{What exactly it is that You pay or receive, can be expressed in utiles. We assume that having more utiles is always the desired outcome, and that this scales linearly.} that depends on something You are uncertain about.
Rejecting an option is making the statement that accepting it is something You don't agree to.
However, it may be that You don't have enough information to make any accept or reject statement about a given option; in that case You're allowed to remain uncommitted.

We'll assume that options~\(\opt\) can be added and multiplied with real numbers, with all the usual properties, so that they live in some real linear space~\(\opts\), called the \define{option space}.
The null option~\(0\) represents the status quo, and we'll assume that Your decision problem is \define{non-trivial} in the sense that \(\opts\neq\set{0}\).
By talking about such abstract options rather than about the more specific cases of gambles \cite{quaeghebeur2015:statement} or quantum measurements \cite{devos2023:indistinguishability,benavoli2016:quantum_2016}, we're able to establish a framework that encompasses both classical and quantum probability.

The set of those options You accept is denoted by \(\A_\acc\) and the set of options You reject by \(\A_\rej\).
Together, these sets form an \define{assessment} \(\A\coloneqq\Adelim{\A_\acc}{\A_\rej}\) that describes Your behaviour regarding options You care to make statements about.
We'll collect all possible assessments in the set~\(\As\coloneqq\powersetof\opts\times\powersetof\opts\).
The options You are uncommitted about --- the \define{unresolved} options --- are given by
\(\A_{\unres}\coloneqq\compof{\A_\acc\cup\A_\rej}\), where `\(\comp\)' denotes the set complement.
We can also identify Your set of \define{indifferent} options \(\A_\indif\coloneqq\A_\acc\cap-\A_\acc\) and Your set of \define{desirable} options \(\A_\des\coloneqq\A_\acc\cap-\A_\rej\):\footnote{The minus sign before a set denotes the Minkowski additive inverse. For example, \(-\A_\acc\coloneqq\set{-\opt\given\opt\in\A_\acc}\).}  You are indifferent about an option when You want to get it \emph{and} give it away, and You find an option desirable when You want it \emph{but} don't want to give it away.

If we define inclusion for assessments by \(\A\subseteq\otherA\ifandonlyif\group{\A_\acc\subseteq\otherA_\acc\text{ and }\A_\rej\subseteq\otherA_\rej}\) for all \(\A,\otherA\in\As\), then the structure \(\structure{\As,\subseteq}\) is a complete lattice with meet and join defined by \(\A\intersection\otherA\coloneqq\Adelim{\A_\acc\intersection\otherA_\acc}{\A_\rej\intersection\otherA_\rej}\) and \(\A\union\otherA\coloneqq\Adelim{\A_\acc\union\otherA_\acc}{\A_\rej\union\otherA_\rej}\), top \(\Astop\coloneqq\Adelim{\opts}{\opts}\) and bottom \(\Adelim{\emptyset}{\emptyset}\).
If \(\A\subseteq\otherA\), we say that \(\A\) is \define{less resolved} than \(\otherA\).

\subsection{Statement models}\label{sec::statement:models}
We'll be concerned with a special subset \(\Ms\subseteq\As\) of assessments, called \define{statement models}.
For an assessment to be called a statement model, four criteria have to be met:
\begin{enumerate}[label={\upshape M\arabic*.},ref={\upshape M\arabic*},series=AR,widest=4,leftmargin=*,itemsep=0pt]
\item\label{axiom:model:background:respected} \(0\in\M_\acc\);\nameit{indifference to status quo}
\item\label{axiom:model:null:not:rejected} \(0\notin\M_\rej\);\nameit{strictness}
\item\label{axiom:model:accepts:convex:cone} \(\M_\acc\) is a convex cone;\nameit{deductive closure}
\item\label{axiom:model:no:limbo} \(\shullof{\M_\rej}-\M_\acc\subseteq\M_\rej\).\nameit{no limbo}
\end{enumerate}
Here, \(\shullof{\M_\rej}\) is the set of all (strictly) positive scalar multiples of the elements of \(\M_\rej\).
We see that \(\structure{\Ms,\subseteq}\) is a \emph{complete meet-semilattice}: \(\Ms\) is closed under taking non-empty infima.
This allows us to define a \define{closure operator}~\(\Mcls\colon\As\to\Ms\cup\set{\Astop}\colon\A\mapsto\Intersection\set{\M\in\Ms\given\A\subseteq\M}\) that maps any assessment to the \emph{least resolved} statement model including it, if any.
This condition, namely \(\Mclsof{\A}\in\Ms\), will be satisfied if and only if \(\posiof{\A_\acc}\cap\A_\rej=\emptyset\), in which case we'll call \(\A\) \define{deductively closable}.
In that case also
\begin{equation}\label{eq:model:closure}
\Mclsof{\A}
=\Adelim{\posiof{\A_\acc}}{\shullof{\A_\rej}\cup\group{\shullof{\A_\rej}-\posiof{\A_\acc}}},
\end{equation}
where \(\posiof{B}\) is the smallest convex cone that includes the set \(B\subseteq\opts\).
\labelcref{axiom:model:background:respected,axiom:model:null:not:rejected,axiom:model:accepts:convex:cone,axiom:model:no:limbo} are the minimal rationality requirements we'll impose on any accept-reject statements You make.
For a thorough discussion of these and all other results and notions mentioned in this \cref{sec::accept:reject}, we refer to \cite{quaeghebeur2015:statement}.

\subsection{About the background}\label{sec::background}
Axiom~\labelcref{axiom:model:background:respected}, or equivalently, \(\zeroM\coloneqq\Adelim{\set{0}}{\emptyset}\subseteq\M\), requires that You should accept --- even be indifferent to --- the \define{status quo}, represented by the null option~\(0\).
\(\zeroM\) is an example of what we'll call a \define{background} (statement) \define{model}: a statement model that You always accept without any real introspection, regardless of any relevant information You might have.
In specific cases, this background model, generically denoted by~\(\bgM\coloneqq\Adelim{\bgM_\acc}{\bgM_\rej}\in\Ms\), may be larger, but we'll always have \(\zeroM\subseteq\bgM\).
We'll then require that the \define{background}~\(\bgM\) should be \define{respected}, which amounts to replacing \labelcref{axiom:model:background:respected} by
\begin{enumerate}[label={\upshape M1*.},ref={\upshape M1*},leftmargin=*,itemsep=0pt]
\item\label{axiom:stronger:background:respected} \(\bgM\subseteq\M\).\nameit{background}
\end{enumerate}
The set of all statement models that respect~\(\bgM\) is denoted by~\(\Msabove{\bgM}\coloneqq\set{\M\in\Ms\given\bgM\subseteq\M}\); for the corresponding closure operator we have that \(\Mclsbgof{\bgM}{\A}=\Mclsof{\bgM\union\A}\) for all \(\A\in\As\).
The background model~\(\bgM\) is the bottom of the complete meet-semilattice \(\structure{\Msabove{\bgM},\subseteq}\), and it's also called the \define{vacuous} statement model.

\subsection{The Desirability-Indifference framework}\label{sec::desirability:indifference}
In the context of this paper, we'll focus on a special type of statement models:
\begin{definition}
A statement model \(\M\in\Ms\) is called a \define{Desirability-Indifference model}, or \define{DI model} for short, if it satisfies the Desirability-Indifference condition:
\begin{equation}\label{eq:DI:condition}
\M_\rej\subseteq-\M_\acc
\text{ and }
\M_\acc=\M_\des\union\M_\indif,
\tag{DI}
\end{equation}
\end{definition}
\noindent which requires that You should only reject options that You want to give away, and that for any option You accept, You should be resolved about wanting or not wanting to give it away.
An equivalent description is given by \(\M_\des=-\M_\rej\) and \(\M_\indif=\M_\acc\setminus-\M_\rej\).
A DI model~\(\M\) can therefore be completely described by specifying its set of desirable options~\(\M_\des\) and its set of indifferent options~\(\M_\indif\), as \(\M=\Adelim{\M_\des\cup\M_\indif}{-\M_\des}\).
We'll denote the set of all DI models by~\(\difMs\), and the set of all DI models that respect a given DI background model~\(\bgM\) by \(\difMsabove{\bgM}\), with \(\bgM\in\difMsabove{\zeroM}\).

For DI models, the rationality criteria \labelcref{axiom:stronger:background:respected,axiom:model:null:not:rejected,axiom:model:accepts:convex:cone,axiom:model:no:limbo} can be rewritten as follows in terms of the sets \(\M_\des\) and \(\M_\indif\):
\begin{theorem}
Consider any background \(\bgM\in\difMsabove{\zeroM}\), then an assessment \(M\) is a DI model that respects the background~\(\bgM\) if and only if
\begin{enumerate}[label={\upshape DI\arabic*.},ref={\upshape DI\arabic*},series=DI,widest=4,leftmargin=*,itemsep=0pt]
\item\label{axiom:DI:background} \(V\subseteq M\);\nameit{background}
\item\label{axiom:DI:zero:not:desirable} \(0\notin\M_\des\);\nameit{strictness}
\item\label{axiom:DI:deductive:closedness} \(\M_\des\) is a convex cone and \(M_\indif\) is a linear space;\nameit{deductive closedness}
\item\label{axiom:DI:compatibility} \(\M_\des+\M_\indif\subseteq\M_\des\).\nameit{compatibility}
\end{enumerate}
\end{theorem}

\subsection{The desirability framework}\label{sec::desirability}
Now is a good moment to realign ourselves with the more familiar approach to dealing with sets of desirable options and sets of indifferent options in the imprecise probabilities literature --- see for instance \cite{cooman2010,decooman2015:coherent:predictive:inference,devos2023:indistinguishability,debock2016:partial:exchangeability} --- and with the generic notation \(\desirset\) for sets of desirable options and \(\indifset\) for sets of indifferent options, rather than \(\M_\des\) and \(\M_\indif\) respectively.
With this change of notation, a reader familiar with that literature  will no doubt recognise \labelcref{axiom:DI:background,axiom:DI:zero:not:desirable,axiom:DI:deductive:closedness,axiom:DI:compatibility} as the typical conditions imposed on compatible sets of desirable and indifferent options.

Something else that's also implicitly done in that literature, and which we'll adhere to here as well, is to focus solely on Your desirability assessments~\(\desirset\), and keep all aspects of indifference~\(\indifset\) in the background model \(\bgM\).
In this co-called \emph{Desirability framework}, Your desirability assessment \(\A=\Adelim{\desirset\cup\set{0}}{-\desirset}\) is then combined with a background \(\bgM=\Adelim{\bgM_\des\cup\indifset}{-\bgM_\des}\in\difMsabove{\zeroM}\), where \(\indifset\) is some given linear space, and \(\bgM_\des\) is some given convex cone for which \(0\notin\bgM_\des\).
The latter defines a strict vector ordering~\(\optgt\) on \(\opts\) by
\[
\opt\optgt\altopt
\ifandonlyif
\opt-\altopt\in\bgM_\des,
\text{ for all \(\opt,\altopt\in\opts\)},
\]
which we'll call the \define{background ordering}.
We'll also take \(\opt\optgeq0\) to mean that `\(\opt\optgt0\) or \(\opt=0\)'.
Observe that, conversely, the background ordering \(\optgt\) also uniquely determines the background cone \(\bgM_\des\), because
\[
\bgM_\des
=\set{\opt\in\opts\given\opt\optgt0}
\eqqcolon\posopts.
\]
That this background \(\bgM=\Adelim{\posopts\cup\indifset}{-\posopts}\) should be a DI model now only amounts to imposing the compatibility condition~\labelcref{axiom:DI:compatibility}:
\begin{equation}\label{eq:background:compatibility}
\posopts+\indifset\subseteq\posopts,
\end{equation}
which, again, the reader conversant in the relevant literature will no doubt recognise.
If we also assume, as is typically done in the literature, that Your set of desirable options~\(\desirset\) is \define{coherent}, so satisfies the conditions
\begin{enumerate}[label={\upshape D\arabic*.},ref={\upshape D\arabic*},series=D,widest=3,leftmargin=*,itemsep=0pt]
\item\label{axiom:D:background} \(\posopts\subseteq\desirset\);\nameit{background}
\item\label{axiom:D:zero:not:desirable} \(0\notin\desirset\);\nameit{strictness}
\item\label{axiom:D:deductive:closedness} \(\desirset\) is a convex cone,\nameit{deductive closedness}
\end{enumerate}
then, since \(\bgM\cup\A=\Adelim{\desirset\cup\indifset}{-\desirset}\) and
\begin{align*}
\posiof{\desirset\cup\indifset}
&=\posiof{\desirset}\cup\posiof{\indifset}\cup\group{\posiof{\desirset}+\posiof{\indifset}}\\
&=\desirset\cup\indifset\cup\group{\desirset+\indifset}
=\indifset\cup\group{\desirset+\indifset},
\end{align*}
it becomes an easy exercise to find the least resolved model~\(\Mclsof{\bgM\cup\A}\) that includes Your set desirable options~\(\desirset\) as well as the background~\(\bgM\), if it exists.
The results in \cref{sec::statement:models} tell us that the least resolved model exists if the \(\bgM\cup\A\) is deductively closable, which is the case if and only if \(\posiof{\desirset\cup\indifset}\cap-\desirset=\emptyset\), so if and only if
\begin{equation}\label{eq:D:compatibility}
\desirset\cap\indifset=\emptyset,
\text{ or equivalently, }
0\notin\desirset+\indifset.
\end{equation}
The least resolved model we're after, is then given by
\begin{equation}\label{eq:least:resolved:model}
\Mclsof{\bgM\cup\A}
=\Adelim{\group{\desirset+\indifset}\cup\indifset}{-\group{\desirset+\indifset}},
\end{equation}
and this is the DI model with set of desirable options \(\desirset+\indifset\) and set of indifferent options \(\indifset\).

\section{Events to condition on}\label{sec::events}
What we're especially interested in, is the following situation: after You've stated Your preferences by means of some initial DI model \(\M\coloneqq\Adelim{\group{\desirset+\indifset}\cup\indifset}{-\group{\desirset+\indifset}}\), You gain new information: some event \(\eventopt\) has occurred.
How to change Your initial preferences in the presence of this newly acquired information?

For the sake of simplicity, we're going to assume that \(\indifset=\set{0}\), so \(\M=\Adelim{\desirset\cup\set{0}}{-\desirset}\in\difMsabove{\bgM}\), with initial background model \(\bgM\coloneqq\Adelim{\posopts\cup\set{0}}{-\posopts}\).
This, incidentally, is no real restriction, as we can always make sure of it by moving to the representation of \(\desirset+\indifset\) in the quotient space \(\quotientopts\), where the null option corresponds to \(\indifset\); see for instance the discussions in \cite{decooman2015:coherent:predictive:inference,debock2016:partial:exchangeability}.

\subsection{Events and their properties}
We'll consider an \define{event} \(\eventopt\) to be a special type of option that can be used to call off any option \(\opt\in\opts\), resulting in the \define{called-off option} \(\calledoff{\opt}\).
We collect all events in the set \(\eventopts\) and call \(\calledoff{\opts}\coloneqq\set{\calledoff{\opt}\given\opt\in\opts}\) the \define{called-off space}.

As before for options, we'll not specify {\itshape in concreto} the exact form of an event, nor what the exact result of calling off an option looks like, but we'll keep the discussion as abstract as possible by merely imposing a number of properties, \labelcref{axiom:event:linear,axiom:event:idempotent,axiom:event:monotone,axiom:event:unit:event,axiom:event:null:event,axiom:event:archimedeanity,axiom:event:ordering,axiom:event:complement}, on the set of events~\(\eventopts\) and the calling-off operation~\(\ast\colon\eventopts\times\opts\to\opts\colon(\eventopt,\opt)\mapsto\calledoff{\opt}\).
In this way, we can keep the discussion general enough to encompass conditioning on events in \emph{classical probability} and on observations in \emph{quantum probability}, amongst others.
These special instantiations of our general framework will be briefly discussed in \cref{sec::instances}, which may also serve a source of intuition for the more abstract discussion here.
\begin{enumerate}[label={\upshape E\arabic*.},ref={\upshape E\arabic*},series=events,widest=3,leftmargin=*,itemsep=0pt]
\item\label{axiom:event:linear} \(\calledoff{(\opt+\lambda\altopt)}=\calledoff{\opt}+\lambda\group{\calledoff{\altopt}}\) for \(\eventopt\in\eventopts\), \(\opt,\altopt\in\opts\) and \(\lambda\in\reals\);\nameit{linearity}
\item\label{axiom:event:idempotent}\(\calledoff{\opt}=\calledoff{(\calledoff{\opt})}\) for all \(\eventopt\in\eventopts\) and all \(\opt\in\opts\);\nameit{idempotency}
\item\label{axiom:event:monotone} \(\opt\optgeq0\then\calledoff{\opt}\optgeq0\) for all \(\eventopt\in\eventopts\) and all \(\opt\in\opts\).\nameit{monotonicity}
\end{enumerate}
By \labelcref{axiom:event:linear,axiom:event:idempotent}, the calling-off operation \(\calledoff{\bolleke}\) is a linear projection, whose range is given by the called-off space \(\calledoff{\opts}\) and whose kernel is given by
\begin{equation}\label{eq:indifset:event}
\eventindifset
\coloneqq\set{\opt\in\opts\given\calledoff{\opt}=0}.
\end{equation}
We'll furthermore assume that there's some \define{unit event} \(\uniteventopt\in\eventopts\) such that
\begin{enumerate}[resume*=events,widest=4]
\item\label{axiom:event:unit:event}  \(\uniteventopt\optgeq0\) and \(\uniteventopt\ast\opt=\opt\) and \(\eventopt\ast\uniteventopt=\eventopt\) for all \(\opt\in\opts\) and all \(\eventopt\in\eventopts\),\nameit{unit event}
\end{enumerate}
and that there's some \define{null event} \(\nulleventopt\in\eventopts\) such that
\begin{enumerate}[resume*=events,widest=5]
\item\label{axiom:event:null:event}  \(\nulleventopt=0\) and \(\nulleventopt\ast\opt=0\) for all \(\opt\in\opts\).\nameit{null event}
\end{enumerate}
The unit event is assumed to be special in that it allows for the following Archimedeanity property:
\begin{enumerate}[resume*=events,widest=6]
\item\label{axiom:event:archimedeanity} for any \(\opt\in\opts\) there's some \(\alpha\in\reals\) such that \(\opt+\alpha\uniteventopt\optgeq0\).\nameit{Archimedeanity}
\end{enumerate}
It follows readily from these assumptions that
\begin{multline}\label{eq:event:extra:properties}
\eventopt\ast\eventopt=\eventopt
\text{ and }
\eventopt\ast\nulleventopt=0\\
\text{ and }
\nulleventopt\optleq\eventopt
\text{ and }
\nulleventopt\optlt\uniteventopt,
\text{ for all \(\eventopt\in\eventopts\)}
\end{multline}
and that
\begin{multline}\label{eq:event:bijection}
\group{\forall\opt\in\opts}
\group{\eventopt[1]\ast\opt=\eventopt[2]\ast\opt}
\ifandonlyif\eventopt[1]=\eventopt[2],\\
\text{ for all \(\eventopt[1],\eventopt[2]\in\eventopts\)},
\end{multline}
which tells us that the calling-off operations \(\calledoff{\bolleke}\) are in a one-to-one relation with the events~\(\eventopt\).
We define the \define{event ordering}~\(\posetleq\) on the set of events~\(\eventopts\) as follows:
\begin{multline}\label{eq:event:ordering}
\eventopt[1]\posetleq\eventopt[2]
\ifandonlyif
\group{\forall\opt\in\opts}\big(\eventopt[1]\ast\opt=\eventopt[1]\ast(\eventopt[2]\ast\opt)\\
=\eventopt[2]\ast(\eventopt[1]\ast\opt)\big),
\text{ for all \(\eventopt[1],\eventopt[2]\in\eventopts\)}.
\end{multline}
Since the relation \(\posetleq\) is reflexive, antisymmetric and transitive, \(\posetleq\) is a partial order on \(\eventopts\).
It has bottom~\(\nulleventopt\) and top~\(\uniteventopt\).
We now impose a condition that links the event ordering \(\posetleq\) to the background ordering of options~\(\optleq\):
\begin{enumerate}[resume*=events,widest=7]
\item\label{axiom:event:ordering} \(\group{\forall\opt\in\opts}\group{\eventopt[2]\ast\opt=0\then\eventopt[1]\ast\opt\not\optgt0}\then\eventopt[1]\posetleq\eventopt[2]\),\\
for all \(\eventopt[1],\eventopt[2]\in\eventopts\).\nameit{ordering of events}
\end{enumerate}
This assumption makes sure that the event ordering \(\posetleq\) can be related to the kernels of the relevant projections.
\begin{proposition}\label{prop:ordering:equivalent:statements}
For all events \(\eventopt[1],\eventopt[2]\in\eventopts\), the following statements are equivalent:
\begin{enumerate}[label={\upshape(\roman*)},ref={\upshape(\roman*)},widest=iii,leftmargin=*,itemsep=0pt]
\item\label{prop:event:ordering} \(\eventopt[1]\posetleq\eventopt[2]\);
\item\label{prop:event:kernels} \(\eventindifset[2]\subseteq\eventindifset[1]\);
\item\label{prop:event:nonpositivity} \(\group{\forall\opt\in\opts}\group{\calledoff[2]{\opt}=0\then\calledoff[1]{\opt}\not\optgt0}\).
\end{enumerate}
\end{proposition}
\noindent An interesting consequence of \cref{prop:ordering:equivalent:statements} is that the kernel \(\eventindifset\) of an event \(\eventopt\) is in direct correspondence with the event itself.
This result follows directly from the equivalence of statements \labelcref{prop:event:ordering} and \labelcref{prop:event:kernels} in \cref{prop:ordering:equivalent:statements} and the fact that \(\posetleq\) is a partial order:
\begin{multline}\label{eq:event:kernels identify events}
\group{\forall\opt\in\opts}\group{\eventopt[1]\ast\opt=0\ifandonlyif\eventopt[2]\ast\opt=0}\ifandonlyif\eventopt[1]=\eventopt[2],\\
\text{ for all \(\eventopt[1],\eventopt[2]\in\eventopts\)}.
\end{multline}
A further, and final, assumption concerns the existence of a \define{complementation operation} \(\neg\colon\eventopts\to\eventopts\colon\eventopt\mapsto\neg\eventopt\) on events:
\begin{enumerate}[resume*=events,widest=8]
\item\label{axiom:event:complement} for every \(\eventopt\in\eventopts\), there's some \(\neg\eventopt\in\eventopts\) such that
\begin{enumerate}[label={\upshape\alph*.},ref={\upshape\alph*},leftmargin=0pt,itemsep=0pt,widest=b]
\item\label{it:event:complement:sum} \(\eventopt+\neg\eventopt\in\eventopts\) and \(\group{\eventopt+\neg\eventopt}\ast\opt=\opt\) for all \(\opt\in\opts\),
\item\label{it:event:complement:nullify} \(\neg\eventopt\ast\group{\eventopt\ast\opt}=0\) for all \(\opt\in\opts\).
\end{enumerate}
\end{enumerate}
Under these assumptions, it's easy to check that
\begin{multline}\label{eq:event:complement:properties}
\eventopt+\neg\eventopt=\uniteventopt
\text{ and }
\neg(\neg\eventopt)=\eventopt\\
\text{ and }
\eventopt\ast\group{\neg\eventopt}
=\group{\neg\eventopt}\ast\eventopt
=0
\text{ for all \(\eventopt\in\eventopts\)},
\end{multline}
so we find in particular that the \define{complement} \(\neg\eventopt\) of an event \(\eventopt\) is unique.

\ifarXiv
\noindent Finally, we add a condition that allows us to replace the sum of the kernels of multiple calling-off operations to the kernel of a single calling-off operation:
\begin{enumerate}[resume*=events,widest=9]
\item\label{axiom:event:kernel:sum} for all \(\eventopt[1],\eventopt[2]\in\eventopts\), there's some \(\eventopt[1]\sqcap\eventopt[2]\in\eventopts\) such that \(\eventindifset[1]+\eventindifset[2]=\indifset[{\eventopt[1]\sqcap\eventopt[2]}]\).
\end{enumerate}
The conditions \labelcref{axiom:event:linear,axiom:event:idempotent,axiom:event:monotone,axiom:event:unit:event,axiom:event:null:event,axiom:event:archimedeanity,axiom:event:ordering,axiom:event:complement,axiom:event:kernel:sum} are internally consistent: as already mentioned above, we'll uncover in \Cref{sec::instances} two specific instances, namely events in classical probability and projections in quantum probability, where all these assumptions are satisfied.
\else
The conditions \labelcref{axiom:event:linear,axiom:event:idempotent,axiom:event:monotone,axiom:event:unit:event,axiom:event:null:event,axiom:event:archimedeanity,axiom:event:ordering,axiom:event:complement} are internally consistent: as already mentioned above, we'll uncover in \Cref{sec::instances} two specific instances, namely events in classical probability and projections in quantum probability, where all these assumptions are satisfied.
\fi

\subsection{Conditioning on an event}\label{sec::conditioning:event}
The effect of the occurrence of an event~\(\eventopt\) is that You can no longer distinguish between any two options whose called-off options coincide: they've now become indifferent to You.
This leads to a new linear space of indifferent options, given by the kernel \(\eventindifset=\set{\opt\in\opts\given\calledoff{\opt}=0}\) of the calling-off operation~\(\calledoff{\bolleke}\).
It therefore seems perfectly reasonable to represent the new knowledge that the event \(\eventopt\) has occurred by the indifference assessment
\begin{equation}\label{eq:model:for:event}
\eventM
\coloneqq\Adelim{\eventindifset}{\emptyset}\in\difMsabove{\zeroM}.
\end{equation}

Which options should You find desirable after the occurrence of the event~\(\eventopt\)?
We're going to assume, in the spirit of discussions by \citeauthor{finetti1937} \cite{finetti1937,finetti19745}, \citeauthor{williams2007} \cite{williams2007} and \citeauthor{walley2000} \cite{walley2000} that You now find desirable all options whose called-off versions were already desirable before the occurrence of the event, leading to the \define{conditional}, or \define{updated}, set of desirable options
\begin{equation}\label{eq:updated:desirable}
\desirset\altcondon\eventopt
\coloneqq\set{\opt\in\opts\given\calledoff{\opt}\in\desirset}.
\end{equation}
Even though it's guaranteed to satisfy \labelcref{axiom:D:zero:not:desirable,axiom:D:deductive:closedness}, this new set of desirable options~\(\desirset\altcondon\eventopt\) won't necessarily be coherent, because not necessarily \(\posopts\subseteq\desirset\altcondon\eventopt\): the original background ordering~\(\optgt\) may not be compatible with the new information that \(\eventopt\) has occurred.

To see what may go wrong, let's find out what happens when we merge the new information in the shape of \(\eventM=\Adelim{\eventindifset}{\emptyset}\) into the existing background, so when we combine \(\bgM\) with the new background information \(\eventM\) to get \(\bgM\union\Adelim{\eventindifset}{\emptyset}=\Adelim{\posopts\cup\eventindifset}{-\posopts}\).
This assessment is deductively closable only if \(\posopts\cap\eventindifset=\emptyset\), or equivalently, if
\begin{equation}\label{eq:background:compatibility:event}
\opt\optgt0\then\calledoff{\opt}\optgt0
\text{ for all~\(\opt\in\opts\)},
\end{equation}
in which case we'll call the event \(\eventopt\) \define{regular}.
Interestingly, the statement in \cref{eq:background:compatibility:event} implies that \(\eventopt=\uniteventopt\), so the only regular event is \(\uniteventopt\).
In that case, \labelcref{axiom:event:unit:event} guarantees in a rather trivial manner that \(\desirset\altcondon\uniteventopt=\desirset\), \(\indifset[\uniteventopt]=\set{0}\) and \(\M_{\uniteventopt}=\Adelim{\set{0}}{\emptyset}\); conditioning on the regular unit event~\(\uniteventopt\) results in no change at all.

So we see that the more interesting events won't be regular, and since we're only taking into account called-off options, we must then \emph{reduce} the background ordering to its restriction to the called-off space.
This leads to a \emph{revised} background ordering, denoted by \(\eventoptgt\):
\begin{equation}\label{eq:background:ordering:event}
\opt\eventoptgt0\ifandonlyif\group{\opt\in\calledoff{\opts}\text{ and }\opt\optgt0},
\text{ for all~\(\opt\in\opts\)},
\end{equation}
so \(\eventposopts=\posopts\cap\eventopt\ast\opts\), by \cref{eq:updated:desirable}.
This ordering is non-empty if and only if there's some \(\opt\in\opts\) such that \(\calledoff\opt\optgt0\), or in other words, if \(\posopts\cap\calledoff{\opts}\neq\emptyset\); we'll then call the event \(\eventopt\) \define{proper}.
Interestingly, the only improper event is the null event~\(\nulleventopt=0\), for which \(\desirset\altcondon\nulleventopt=\emptyset\), \(\indifset[\nulleventopt]=\opts\) and \(\M_{\nulleventopt}=\Adelim{\opts}{\emptyset}\).
For the regular (so proper) event \(\uniteventopt\), clearly \(\opt\optgt[\uniteventopt]0\ifandonlyif\opt\optgt0\), so the revised background ordering \(\optgt[\uniteventopt]\) is the same as the original one.

Reducing \(\optgt\) to \(\eventoptgt\) makes sure that \cref{eq:background:compatibility:event} now holds trivially, so the revised background assessment \(\Adelim{\eventposopts\cup\eventindifset}{-\eventposopts}\) is deductively closable.
Taking its closure leads to the \emph{revised} background DI model
\begin{equation}\label{eq:background:revised}
\eventbgM
\coloneqq\Adelim{\group{\eventposopts+\eventindifset}\cup\eventindifset}{-\group{\eventposopts+\eventindifset}},
\quad\eventopt\in\eventopts.
\end{equation}
Since, trivially, \(\eventposopts\subseteq\desirset\altcondon\eventopt\) and \(\desirset\altcondon\eventopt+\eventindifset=\desirset\altcondon\eventopt\), we infer from the discussion in \Cref{sec::desirability}, and in particular \cref{eq:least:resolved:model}, that the DI model
\begin{equation}\label{eq:desirability:indifference:revised}
\Adelim{\desirset\altcondon\eventopt\cup\eventindifset}{-\desirset\altcondon\eventopt}\in\difMsabove{\eventbgM},
\quad\eventopt\in\eventopts
\end{equation}
is the \emph{least resolved statement model that combines the updated set of desirable options \(\desirset\altcondon\eventopt\) with the revised background DI model \(\eventbgM\), or equivalently, with the new background information~\(\eventM=\Adelim{\eventindifset}{\emptyset}\).}
This holds in particular for the regular unit event \(\uniteventopt\), which is necessarily proper because we've already shown that \(\uniteventopt\optgt0\) [see \cref{eq:event:extra:properties}] and for which \(\bgM_{\uniteventopt}=\bgM\).

We see that the occurrence of an event \(\eventopt\) leads You to replace Your initial statement model with a new, \define{updated}, one.
This is a form of belief change, and we are thus led to wonder if we can describe it using the ideas and tools introduced by \citeauthor{alchourron1985} (AGM) in \cite{alchourron1985}; see also the general discussion in \cite{gardenfors1988} and the very thorough overview in \cite{rott2001}.

It's not hard to see that we won't generally have that \(\desirset\subseteq\desirset\altcondon\eventopt\), nor that
\begin{equation*}
\bgM\subseteq\eventbgM
\text{ or }
\Adelim{\desirset\cup\set{0}}{-\desirset}\subseteq\Adelim{\desirset\altcondon\eventopt\cup\eventindifset}{-\desirset\altcondon\eventopt}.
\end{equation*}
Taking into account the extra information that an event \(\eventopt\) has occurred therefore doesn't necessarily lead to an increase in resolve: updating isn't necessarily a monotonic operation.
Another way of saying this, is that updating typically can't be seen as belief \emph{expansion}, but should rather be viewed as a form of belief \emph{revision}.

But, as \citeauthor{alchourron1985} in \cite{alchourron1985} and \citeauthor{gardenfors1988} in \cite{gardenfors1988} focused mainly on belief states that are sets of propositions, or that are probability measures, we need a way to generalise their ideas to the more abstract setting where the belief states are statement models.
The tools for this were developed by one of us in a paper that was published quite some time ago \cite{cooman2003a}, and we'll use the next section to briefly summarise the ideas there that are directly relevant to the present discussion.

\section{Belief states and belief change}\label{sec::belief:change}
We'll consider abstract objects, called \define{belief models}, or \define{belief states}, and collect them in a set \(\belmods\).
We assume that the belief states~\(\belmod\) in \(\belmods\) are partially ordered by a binary relation \(\modleq\), and that they constitute a \emph{complete lattice} \(\structure{\belmods,\modleq}\).
Denote its top as \(\belmodtop\) and its bottom as \(\belmodbottom\), its meet as \(\meet\) and its join as \(\join\).

We're particularly interested in a special non-empty subset \(\cohmods\subseteq\belmods\) of belief states, whose elements are called \define{coherent} and are considered to be `more perfect' than the others.
The inherited partial ordering \(\modleq\) on \(\cohmods\) is interpreted as `is more conservative than'.

Crucially, we'll assume that \(\cohmods\) is closed under arbitrary non-empty infima, so for every non-empty subset \(C\subseteq\cohmods\) we assume that \(\inf C\in\cohmods\).
Additionally, we'll assume that the complete meet-semilattice \(\structure{\cohmods,\modleq}\) has no top --- so definitely \(\belmodtop\notin\cohmods\) --- implying that there's a smallest (most conservative) coherent belief state \(\cohmodbottom=\inf\cohmods\) but no largest (least conservative) one.
We'll let the incoherent \(\belmodtop\) represent contradiction.
Of course, the set \(\closedmods\coloneqq\cohmods\cup\set{\belmodtop}\) provided with the partial ordering \(\modleq\) is a complete lattice.
The corresponding triple \(\structure{\belmods,\cohmods,\modleq}\) is then called a \define{belief structure}.

That \(\cohmods\) is closed under arbitrary non-empty infima, leads us to define a closure operator as follows:
\begin{equation*}
\cohmodclosure
\colon\belmods\to\closedmods
\colon\belmod\mapsto\cohmodclosureof{\belmod}\coloneqq\inf\set{\cohmod\in\cohmods\given\belmod\modleq\cohmod},
\end{equation*}
so \(\cohmodclosureof{\belmod}\) is the most conservative coherent belief state that is at least as committal as \(\belmod\) --- if any.
The closure operator implements \emph{conservative inference} in the belief structure \(\structure{\belmods,\cohmods,\modleq}\).
Using this closure~\(\cohmodclosure\), we can now define notions of \define{consistency} and \define{closedness}: a belief state \(\belmod\in\belmods\) is \define{consistent} if \(\cohmodclosureof{\belmod}\neq\belmodtop\), or equivalently, if \(\cohmodclosureof{\belmod}\in\cohmods\), and \define{closed} if \(\cohmodclosureof{\belmod}=\belmod\).
Two belief states \(\belmod[1],\belmod[2]\in\belmods\) are called \define{consistent} if their join \(\belmod[1]\join\belmod[2]\) is.
We see that \(\closedmods\) is the set of all closed belief states, that the coherent belief states in \(\cohmods\) are the ones that are both consistent and closed, and that \(\belmodtop\) represents inconsistency for the closed belief states.

The statement models \(\M\in\Msabove{\zeroM}\) that respect indifference to the status quo, provided with the `is at most as resolved as' ordering \(\subseteq\), can be seen to constitute a special instance of these belief structures, with the correspondences identified in \Cref{table:correspondences}.

\begin{table}[ht]
\begin{tabular}{lcl}
\(\belmods\)
&\(\instantiateas\)
&\(\As\)\\
\(\modleq\)
&\(\instantiateas\)
&\(\subseteq\)\\
\(\belmodbottom\)
&\(\instantiateas\)
&\(\Adelim{\emptyset}{\emptyset}\)\\
\(\belmodtop\)
&\(\instantiateas\)
&\(\Adelim{\opts}{\opts}\)\\
\(\inf\)
&\(\instantiateas\)
&\(\Intersection\)\\
\(\cohmods\)
&\(\instantiateas\)
&\(\Msabove{\zeroM}\)\\
\(\closedmods\)
&\(\instantiateas\)
&\(\closedMsabove{\zeroM}\coloneqq\Msabove{\zeroM}\cup\set{\Adelim{\opts}{\opts}}\)\\
\(\cohmodbottom\)
&\(\instantiateas\)
&\(\zeroM\)\\
\(\cohmodclosureof{}\)
&\(\instantiateas\)
&\(\Mclsof{\zeroM\union\bolleke}\)\\
consistent
&\(\instantiateas\)
&deductively closable\\
coherent
&\(\instantiateas\)
&statement model that respects \(\zeroM\)
\end{tabular}
\caption{Correspondences between belief states and sets of accept-reject statements}
\label{table:correspondences}
\end{table}

When You're in a coherent belief state \(\cohmod\in\cohmods\), new information the form of some belief state \(\belmod\in\belmods\) --- not necessarily coherent --- can cause You to change Your belief state.
\citeauthor{alchourron1985} \cite{alchourron1985} considered three important types of belief change: belief expansion, belief revision and belief contraction.
They proposed a number of axioms for these operations, and studied and discussed their resulting properties.

De Cooman \cite{cooman2003a} argued that the axioms for belief expansion and belief revision can be elegantly translated to the setting of belief structures --- the counterpart of belief contraction is more problematic.
We'll not discuss these abstract axioms here, but propose to postpone listing their more concrete forms until the next section, where we consider belief change for Desirability-Indifference models caused by the occurrence of events.

\section{Conditioning as belief change}\label{sec::conditioning}
As explained above, You start out with a DI model \(\M=\Adelim{\desirset\cup\set{0}}{-\desirset}\) that respects a background \(\bgM=\Adelim{\posopts\cup\set{0}}{-\posopts}\), and You gain new information in the form of the occurrence of an event \(\eventopt\), which we've argued corresponds to new DI background information \(\eventM=\Adelim{\eventindifset}{\emptyset}\).
We have already mentioned above that, due to \labelcref{eq:event:kernels identify events}, the sets of indifferent options \(\eventindifset\) --- and therefore also the statement models \(\eventM=\Adelim{\eventindifset}{\emptyset}\) --- are in a one-to-one correspondence with the events \(\eventopt\).

One way to combine Your statement model with such new information goes via a \define{belief expansion} operator \footnote{The \(\vert\) symbol merely acts as a way to separate the first from the second argument here.}
\begin{equation*}
\expand
\colon\Msabove{\zeroM}\times\As\to\As
\colon(\M,\A)\mapsto\expandof{\M\given\A},
\end{equation*}
De Cooman \cite{cooman2003a} has argued that the AGM axioms for belief expansion \cite{alchourron1985,gardenfors1988,rott2001} can be translated directly to the abstract setting of belief structures.
His results lead us to conclude that the resulting expansion operator \(\expand\) must then be uniquely determined by the resulting model closure, in the sense that \(\expandof{\M\given\A}=\Mclsbgof{\zeroM}{\M\union\A}\).
In our conditioning setting, this leads to \(\expandof{\M\given\eventM}=\Mclsbgof{\zeroM}{\M\union\eventM}\), where \(\M\union\eventM=\Adelim{\desirset\cup\eventindifset}{-\desirset}\).
Since \(\posiof{\desirset\cup\eventindifset}=\group{\desirset+\eventindifset}\cup\eventindifset\), we find after a few algebraic manipulations that \(\M\union\eventM\) is deductively closable if and only if \(\desirset\cap\eventindifset=\emptyset\), which as we've already argued in \Cref{sec::events} will only be the case if \(\eventopt\) is regular, and therefore only if \(\eventopt=\uniteventopt\).
Recalling that \(\indifset[\uniteventopt]=\set{0}\) and therefore \(\M_{\uniteventopt}=\Adelim{\set{0}}{\emptyset}\), we find that \(\Mclsbgof{\zeroM}{\M\union\eventM}=\Mclsbgof{\zeroM}{\M}=\M\).
Hence,
\begin{equation}\label{eq:expand}
\expandof{\M\given\eventM}\\
=\begin{cases}
\M=\Adelim{\desirset\cup\set{0}}{-\desirset}
&\text{if \(\eventopt=\uniteventopt\)}\\
\Adelim{\opts}{\opts}
&\text{otherwise}.
\end{cases}
\end{equation}
So, unless \(\eventopt=\uniteventopt\), expanding \(\M\) with \(\eventM\) will lead to  inconsistency; this is because \(\desirset\) includes the original background cone \(\posopts\), which, as we've already mentioned in \Cref{sec::events} will only be consistent with the indifferences in \(\eventindifset\) provided that \(\eventindifset\cap\posopts=\emptyset\), which only happens for the uniquely regular event \(\eventopt=\uniteventopt\).

We're therefore led to consider a different type of belief change, based on a \define{belief revision} operator
\begin{equation*}
\revise
\colon\Msabove{\zeroM}\times\As\to\As
\colon(\M,\A)\mapsto\reviseof{\M\given\A}.
\end{equation*}
De Cooman \cite{cooman2003a} has argued that the AGM axioms for belief revision \cite{alchourron1985,gardenfors1988} can be transported to the abstract setting of belief structures.
His revision axioms, which correspond one to one with the AGM revision postulates \(\mathrm{K}^*1\)--\(\mathrm{K}^*8\) in \cite{gardenfors1988}, are the following, when instantiated in the present context of statement models:
\begin{enumerate}[label={\upshape BR\arabic*.},ref={\upshape BR\arabic*},series=BR,widest=8,leftmargin=*,itemsep=0pt]
\item\label{axiom:revision:1} \(\reviseof{\M\given\A}\in\closedMsabove{\zeroM}\) for \(\M\) in \(\Msabove{\zeroM}\) and \(\A\) in \(\As\);
\item\label{axiom:revision:2} \(\A\subseteq\reviseof{\M\given\A}\) for \(\M\) in \(\Msabove{\zeroM}\) and \(\A\) in \(\As\);
\item\label{axiom:revision:3} \(\reviseof{\M\given\A}\subseteq\expandof{\M\given\A}\) for \(\M\) in \(\Msabove{\zeroM}\) and \(\A\) in \(\As\);
\item\label{axiom:revision:4}\(\expandof{\M\given\A}\subseteq\reviseof{\M\given\A}\) for  \(\M\) in \(\Msabove{\zeroM}\) and \(\A\) in \(\As\) such that \(\A\) and \(\M\) are consistent;
\item\label{axiom:revision:5} \(\reviseof{\M\given\A}\) is inconsistent if and only if \(A\) is inconsistent, for \(\M\) in \(\Msabove{\zeroM}\) and \(\A\) in \(\As\);
\item\label{axiom:revision:6} \(\reviseof{\M\given\A}=\reviseof{\M\given\Mclsbgof{\zeroM}{\A}}\) for \(\M\) in \(\Msabove{\zeroM}\) and \(\A\) in \(\As\);
\item\label{axiom:revision:7} \(\reviseof{\M\given\A_1\union\A_2}\subseteq\expandof{\reviseof{\M\given\A_1}\given\A_2}\) for \(\M\) in \(\Msabove{\zeroM}\) and \(A_1,\A_2\) in \(\As\);
\item\label{axiom:revision:8} \(\expandof{\reviseof{\M\given\A_1}\given\A_2}\subseteq\reviseof{\M\given\A_1\union\A_2}\) for \(\M\) in \(\Msabove{\zeroM}\) and \(A_1,\A_2\) in \(\As\) such that \(\reviseof{\M\given\A_1}\) and \(A_2\) are consistent.
\end{enumerate}
Belief revision is necessary when expansion leads to inconsistency [see \labelcref{axiom:revision:3,axiom:revision:4}], and works essentially by preserving \(\A\) [see \labelcref{axiom:revision:2}] and reducing \(\M\).

Inspired by the discussion in \Cref{sec::conditioning:event}, and in particular \cref{eq:desirability:indifference:revised}, we propose the following form for a belief revision operator, whose first argument we'll restrict to DI models \(\M\in\difMsabove{\zeroM}\) of the form \(\M=\Adelim{\desirset\cup\set{0}}{-\desirset}\) where \(\desirset\) is a coherent set of desirable options, and whose second argument we'll restrict to the DI models \(\eventM=\Adelim{\eventindifset}{\emptyset}\), which are in a one-to-one correspondence with the events \(\eventopt\in\eventopts\):
\begin{multline}\label{eq:revision:operator}
\reviseof{\M\given\eventM}
\coloneqq\Adelim{\desirset\altcondon\eventopt\cup\eventindifset}{-\desirset\altcondon\eventopt}\in\difMsabove{\eventbgM},\\
\text{ for all \(\M\in\difMsabove{\zeroM}\) and all \(\eventopt\in\eventopts\)}.
\end{multline}
Remark that we never get to an inconsistency, because, as explained in the previous section, we've allowed ourselves to change the background model as a result of the conditioning operation.

\begin{proposition}\label{prop:revision}
When we restrict the belief revision operator \(\reviseof{\M\given\eventM}\) as introduced in \cref{eq:revision:operator} to DI models \(\M\in\difMsabove{\zeroM}\) of the form \(\M=\Adelim{\desirset\cup\set{0}}{-\desirset}\) where \(\desirset\) is any coherent set of desirable options, and to DI models \(\eventM=\Adelim{\eventindifset}{\emptyset}\) corresponding to events \(\eventopt\in\eventopts\), then it satisfies the correspondingly restricted \labelcref{axiom:revision:1,axiom:revision:2,axiom:revision:3,axiom:revision:4,axiom:revision:5,axiom:revision:6,axiom:revision:7,axiom:revision:8}.
\end{proposition}

\ifarXiv
\noindent Observe, by the way, that by invoking \labelcref{axiom:revision:6,axiom:event:kernel:sum}, we infer from \cref{eq:revision:operator} that
\begin{multline*}
\reviseof{\M\given\indexedeventM[1]\cup\indexedeventM[2]}
=\reviseof{\M\given\M_{\eventopt[1]\sqcap\eventopt[2]}}\\
\text{with \(\M_{\eventopt[1]\sqcap\eventopt[2]}=\Adelim{\eventindifset[1]+\eventindifset[2]}{\emptyset}\).}
\end{multline*}
\fi

\citeauthor{alchourron1985} \cite{alchourron1985,gardenfors1988} also discuss the notion of \emph{belief contraction}, which in our context can be made to correspond to the action of a \define{belief contraction} operator
\begin{equation*}
\contract
\colon\Msabove{\zeroM}\times\As\to\As
\colon(\M,\A)\mapsto\contractof{\M\given\A},
\end{equation*}
for which the following direct counterparts of the AGM contraction postulates could be formulated as follows --- the correspondence is again one to one with the AGM contraction postulates \(\mathrm{K}^-1\)--\(\mathrm{K}^-8\) in \cite{gardenfors1988}:
\begin{enumerate}[label={\upshape BC\arabic*.},ref={\upshape BC\arabic*}, series=BC,widest=8,leftmargin=*,itemsep=0pt]
\item\label{axiom:contraction:1} \(\contractof{\M\given\A}\in\closedMsabove{\zeroM}\) for \(M\) in \(\Msabove{\zeroM}\) and \(\A\in\As\);
\item\label{axiom:contraction:2} \(\contractof{\M\given\A}\subseteq\M\) for \(\M\) in \(\Msabove{\zeroM}\) and \(\A\in\As\);
\item\label{axiom:contraction:3} \(\contractof{\M\given\A}=\M\) for \(\M\) in \(\Msabove{\zeroM}\) and \(\A\in\As\) such that \(\neg\A\) and \(\M\) are consistent;
\item\label{axiom:contraction:4} if \(\A\subseteq\contractof{\M\given\A}\) then \(\neg\A\) is inconsistent, for \(\M\) in \(\Msabove{\zeroM}\) and \(\A\in\As\);
\item\label{axiom:contraction:5} if \(A\subseteq\M\) then \(\M\subseteq\expandof{\contractof{\M\given\A}\given\A}\), for \(\M\) in \(\Msabove{\zeroM}\) and \(\A\in\As\);
\item\label{axiom:contraction:6} \(\contractof{\M\given\A}=\contractof{\M\given\Mclsbgof{\zeroM}{\A}}\) for \(\M\) in \(\Msabove{\zeroM}\) and \(\A\in\As\);
\item\label{axiom:contraction:7} \(\contractof{\M\given\A_1}\intersection\contractof{\M\given\A_2}\subseteq\contractof{\M\given\A_1\union\A_2}\) for \(\M\) in \(\Msabove{\zeroM}\) and \(\A_1,\A_2\in\As\);
\item\label{axiom:contraction:8} \(\contractof{\M\given\A_1\union\A_2}\subseteq\contractof{\M\given\A_1}\) for \(\M\) in \(\Msabove{\zeroM}\) and \(\A_1,\A_2\in\As\) such that \(\neg\A_1\) and \(\contractof{\M\given\A_1\union\A_2}\) are consistent.
\end{enumerate}
Belief contraction aims at removing the consequences of an assessment \(\A\) from a statement model~\(\M\).
For the axioms \labelcref{axiom:contraction:3,axiom:contraction:4,axiom:contraction:8} to make sense at all, we need to be able to introduce a negation operator \(\neg\) for accept-reject statement sets, which is far from obvious in the present context, where such statement sets needs't be propositional.
It's for this reason that De Cooman refrained from discussing belief contraction in \cite{cooman2003a}.

But when we restrict ourselves in the current context to belief contraction with statement models \(\eventM\), which are related to the occurrence of events \(\eventopt\), contraction becomes a more reasonable proposition: for events \(\eventopt\) we have complemented events \(\neg\eventopt\), so we'll define the \define{negated statement model}
\begin{equation}\label{eq:negation}
\neg\eventM
\coloneqq\eventM[\neg\eventopt]
=\Adelim{\indifset[\neg\eventopt]}{\emptyset}
\end{equation}
as the one associated with the complemented event \(\neg\eventopt\).

We propose the following form for a belief contraction operator, whose first argument we'll restrict to DI models \(\M\in\difMsabove{\zeroM}\) of the form \(\M=\Adelim{\desirset\cup\set{0}}{-\desirset}\) and whose second argument we'll restrict to the DI models \(\eventM=\Adelim{\eventindifset}{\emptyset}\), which are in a one-to-one correspondence with the events \(\eventopt\in\eventopts\):
\begin{multline}\label{eq:contraction:operator}
\contractof{\M\given\eventM}
\coloneqq\M\intersection\reviseof{\M\given\neg\eventM},\\
\text{ for all \(\M\in\difMsabove{\zeroM}\) and all \(\eventopt\in\eventopts\)},
\end{multline}
leading to
\begin{multline}\label{eq:contraction:operator:too}
\contractof{\M\given\eventM}
=\Adelim{(\desirset\cup\set{0})\cap(\desirset\altcondon\neg\eventopt\cup\indifset[\neg\eventopt])}{-\group{\desirset\cap\desirset\altcondon\neg\eventopt}},\\
\text{ for all \(\M\in\difMsabove{\zeroM}\) and all \(\eventopt\in\eventopts\)}.
\end{multline}
For the statement model \(\eventM\) where \(\eventopt=\uniteventopt\), this contraction would result in removing the background \(\zeroM\) from the statement model \(\M\), leading to a violation of \labelcref{axiom:contraction:1}.
To avoid this, we'll restrict the DI models \(\eventM\) to those corresponding to non-regular events \(\eventopt\neq\uniteventopt\).

\begin{proposition}\label{prop:contraction}
When we restrict the belief contraction operator \(\contractof{\M\given\eventM}\) as introduced in \cref{eq:contraction:operator} to DI models \(\M\in\difMsabove{\zeroM}\) of the form \(\M=\Adelim{\desirset\cup\set{0}}{-\desirset}\), where \(\desirset\) is any coherent set of desirable options, and to DI models \(\eventM=\Adelim{\eventindifset}{\emptyset}\) corresponding to non-regular events \(\eventopt\in\eventopts\setminus\set{\uniteventopt}\), then it satisfies the correspondingly restricted \labelcref{axiom:contraction:1,axiom:contraction:2,axiom:contraction:3,axiom:contraction:4,axiom:contraction:5,axiom:contraction:6}.
\end{proposition}
\noindent Interestingly, we have a counterexample for \labelcref{axiom:contraction:7} in the case of classical probabilistic inference, which can also be used for quantum probabilistic inference.
It would be interesting, but beyond the scope of the present paper, to check whether alternative versions of \labelcref{axiom:contraction:7} (and \labelcref{axiom:contraction:8}), as they are for instance formulated in a monograph by Rott \cite{rott2001}, would hold in our context.

To finalise the discussion, we mention that also in our present context, as in the propositional context covered by the AGM framework \cite{alchourron1985,gardenfors1988}, versions of Levi's and Harper's identities hold; our version of Harper's identity actually holds by definition.

\begin{proposition}[Harper's identity]\label{prop:identity:harper}
\(M\in\difMsabove{\zeroM}\) of the type \(\M=\Adelim{\desirset\cup\set{0}}{-\desirset}\) and all \(\eventopt\in\eventopts\),
\begin{equation*}
\contractof{\M\given\eventM}
=\M\intersection\reviseof{\M\given\neg\eventM}.
\end{equation*}
\end{proposition}

\begin{proposition}[Levi's identity]\label{prop:identity:levi}
For all \(M\in\difMsabove{\zeroM}\) of the type \(\M=\Adelim{\desirset\cup\set{0}}{-\desirset}\) and all \(\eventopt\in\eventopts\),
\begin{equation*}
\reviseof{\M\given\eventM}
=\expandof{\contractof{\M\given\neg\eventM}\given\eventM}.
\end{equation*}
\end{proposition}

\section{Interesting special instances}\label{sec::instances}
Let's now identify two special instances of the general abstract option and event framework developed above.

\subsection{Classical probabilistic inference}
In a decision-theoretic context related to classical probabilistic reasoning, we consider a variable~\(\varstate\) that may assume values in some non-empty set \(\states\), but whose actual value is unknown to You.
With any bounded map \(\gbl\colon\states\to\reals\), called \define{gamble}, there corresponds an uncertain reward \(\gbl(\varstate)\), expressed in units of some linear utility scale.
The set of all gambles, denoted by \(\gambles(\states)\), constitutes a real linear space under pointwise addition and pointwise scalar multiplication with real numbers.

So, we take as our option space the set \(\opts\instantiateas\gambles(\states)\) of all gambles, which You can express preferences between.
For the background ordering \(\optgt\), we take the \define{weak (strict) preference ordering} \(\weakgt\) defined by
\begin{equation*}
\gbl\weakgt0
\ifandonlyif\underset{\textrm{this defines }\gbl\weakgeq0}{\underbrace{\group{\forall\state\in\states}\gbl(\state)\geq0}}
\text{ and }\gbl\neq0.
\end{equation*}

Here, \emph{events} are subsets~\(\event\) of the possibility space~\(\states\), and they can (and will) be identified with special indicator gambles \(\indevent\) that assume the value \(1\) on \(\event\) and \(0\) elsewhere; so here, \(\eventopts\instantiateas\set{\indevent\given\event\subseteq\states}\).
The \emph{unit event} corresponds to the constant gamble \(1=\indof{\states}\) and the \emph{null event} to the constant gamble \(0=\indof{\emptyset}\).
For the \emph{called-off gambles}, we have that \(\indevent\ast\gbl\instantiateas\indevent\gbl\), and \emph{complementation} corresponds to \(\neg\indevent\instantiateas1-\indevent\).
It's a trivial exercise to show that the assumptions \labelcref{axiom:event:linear,axiom:event:idempotent,axiom:event:monotone,axiom:event:unit:event,axiom:event:null:event,axiom:event:archimedeanity,axiom:event:ordering,axiom:event:complement} are satisfied.
The event ordering \(\posetleq\) corresponds to set inclusion: \(\indevent[1]\posetleq\indevent[2]\ifandonlyif\event[1]\subseteq\event[2]\).
\ifarXiv
\labelcref{axiom:event:kernel:sum} is also satisfied, with \(\indevent[1]\meet\indevent[2]\instantiateas\indof{\event[1]\cap\event[2]}\), as we show explicitly in the Appendix.
\fi
The only \emph{regular event} is the unit event \(1=\indof{\states}\), and the \emph{proper events} \(\indevent\) correspond to the non-empty subsets \(\event\neq\emptyset\) of \(\states\).

Updating a coherent set of desirable gambles \(\desirset\) with a non-empty, and therefore proper, event \(\event\neq\emptyset\) to get to \(\desirset\altcondon\event\), is a well-established operation that leads to a generalisation of \emph{Bayes's rule} for conditioning in probability theory; see for instance \cite{walley1991,walley2000,williams1975,troffaes2013:lp}.

\subsection{Quantum probabilistic inference}
Consider, in a decision-theoretic context related to quantum probabilistic reasoning, a quantum system whose unknown state \(\uket\) lives in a finite-dimensional state space~\(\hilbertspace\), which is a complex Hilbert space whose dimension we'll denote by \(n\).
Options in this context are the Hermitian operators \(\measurement{A}\) on~\(\hilbertspace\) corresponding to \define{measurements} on the system, where the outcome of a measurement~\(\measurement{A}\) is interpreted as an uncertain reward, expressed in units of some linear utility scale, which You can get if you accept the measurement.
The set of all such Hermitian operators~\(\measurement{A}\) constitutes an \(n^2\)-dimensional real linear space~\(\measurements\), and this is Your option space for this quantum decision problem.
We denote by \(\spectrum{A}\) the set of all (real) eigenvalues ---  possible outcomes --- of the measurement~\(\measurement{A}\).
For the background ordering, we take the strict vector ordering associated with \define{positive semidefiniteness}, defined by
\begin{equation*}
\measurement{A}\weakgt0
\ifandonlyif\underset{\textrm{this defines }\measurement{A}\weakgeq0}{\underbrace{\min\spectrum{A}\geq0}}
\text{ and }\measurement{A}\neq0.
\end{equation*}

Here, any \emph{event} corresponds to a subspace~\(\subspace\) of the state space \(\hilbertspace\).
It can be identified with a special measurement \(\projector=\projector\projector\), the (linear and orthogonal) projection operator onto the subspace~\(\subspace\), with eigenvalue \(1\) associated with the eigenspace \(\subspace\) and eigenvalue \(0\) associated with its orthogonal complement \(\subspace^\perp\).
This tells us that \(\eventopts\instantiateas\set{\projector\given\subspace\text{ is a subspace of }\hilbertspace}\).
The \emph{unit event} corresponds to the identity measurement \(\identity=\projectoron{\hilbertspace}\) and the \emph{null event} to the zero measurement \(\zero=\projectoron{\set{0}}\).
For the \emph{called-off measurements}, we have that \(\projector\ast\measurement{A}\instantiateas\projector\measurement{A}\projector\), and \emph{complementation} corresponds to \(\neg\projector\instantiateas\identity-\projector=\projectoron{\subspace^\perp}\).
It's a straightforward exercise to show that \labelcref{axiom:event:linear,axiom:event:idempotent,axiom:event:monotone,axiom:event:unit:event,axiom:event:null:event,axiom:event:archimedeanity} and \labelcref{axiom:event:complement} hold.
A proof for \labelcref{axiom:event:ordering} is given in the Appendix.
The event ordering \(\posetleq\) corresponds to the inclusion ordering of the subspaces:
\begin{align*}
\projector[1]\posetleq\projector[2]
&\ifandonlyif\projector[1]\projector[2]=\projector[2]\projector[1]=\projector[1]\\
&\ifandonlyif\subspace[1]\subseteq\subspace[2].
\end{align*}
\ifarXiv
\labelcref{axiom:event:kernel:sum} is also satisfied, with \(\projector[1]\meet\projector[2]\instantiateas\projectoron{\subspace[1]\cap\subspace[2]}\), as we show explicitly in the Appendix.
\fi
The only regular event is the identity measurement \(\identity=\projectoron{\hilbertspace}\), and the proper events \(\projector\) correspond to the non-null subspaces \(\subspace\neq\set{0}\) of \(\hilbertspace\).

Updating a coherent set of desirable measurements \(\desirset\) with a non-null, and therefore proper, event \(\projector\neq\zero\) to get to \(\desirset\altcondon\subspace\), is an operation that leads to a generalisation of \emph{Lüders' conditioning rule} \cite{devos2025:isipta}.

\section{Conclusion}\label{sec::conclusion}
We were able to identify an abstract option and event structure that allows us to look at conditioning sets of desirable options in terms of belief change operators that are consistent with the AGM framework.
This abstract account is general enough to cover conditioning  both in classical and quantum probability settings.

We've formulated our arguments in the Desirability-Indifference framework, where the background is a DI model.
Recent work \cite{devos2025:isipta}, however, seems to suggest that it might be possible to find even more interesting results by allowing for (somewhat) more general background models, leading us to consider moving to the Accept-Desirability framework \cite{quaeghebeur2015:statement} for future work in this area.
Also, more work is needed to fully relate our work to variants of the AGM framework that have been proposed in the literature, as they are for instance summarised in Rott's monograph \cite{rott2001}.

\appendix

\section{Proofs}

\begin{proof}[Proof of \cref{eq:D:compatibility}]
Since \(\posiof{\desirset\cup\indifset}=\indifset\cup\group{\desirset+\indifset}\), we find that \(\posiof{\desirset\cup\indifset}\cap-\desirset=\emptyset\) if and only if \(0\notin\desirset+\indifset+\desirset=\desirset+\indifset\), or equivalently, \(\desirset\cap\indifset=\emptyset\).
\end{proof}

\begin{proof}[Proof of \cref{eq:least:resolved:model}]
Since we proved above that \(\desirset\cap\indifset=\emptyset\) implies deductive closability, we find, using \cref{eq:model:closure}, that for the accept part of \(\M\coloneqq\Mclsof{\bgM\cup\A}\), \(\M_\acc=\posiof{\desirset\cup\indifset}=\indifset\cup\group{\desirset+\indifset}\) and that for its reject part, \(\M_\rej=-\group{\desirset\cup\group{\desirset+\group{\indifset\cup\group{\desirset+\indifset}}}}=-\group{\desirset\cup\group{\desirset+\indifset}\cup\group{\desirset+\desirset+\indifset}}=-\group{\desirset+\indifset}\).
\end{proof}

\begin{proof}[Proof of \cref{eq:event:extra:properties}]
Consider any \(\eventopt\in\eventopts\).
That \(\eventopt\ast\eventopt=\eventopt\) follows from \labelcref{axiom:event:idempotent}, by letting \(\opt\instantiateas\uniteventopt\) and recalling that \(\eventopt\ast\uniteventopt=\eventopt\), by \labelcref{axiom:event:unit:event}.
That \(\eventopt\ast\nulleventopt=0\) follows from \labelcref{axiom:event:linear}, by letting \(\altopt\instantiateas\opt\) and \(\lambda\instantiateas-1\).
That \(\nulleventopt\optleq\eventopt\) follows from \labelcref{axiom:event:monotone} with \(\opt\instantiateas\uniteventopt\), taking into account \labelcref{axiom:event:unit:event}.
That, finally, \(\nulleventopt\optlt\uniteventopt\) follows from \labelcref{axiom:event:unit:event,axiom:event:null:event} and our general non-triviality assumption that \(\opts\neq\set{0}\).
\end{proof}

\begin{proof}[Proof of \cref{eq:event:bijection}]
It's clear that only the direct implication needs any attention.
Letting \(\opt\instantiateas\uniteventopt\) on the left-hand side, and applying \labelcref{axiom:event:unit:event} leads directly to the right-hand side of the desired implication.
\end{proof}

\begin{proof}[Proof that \(\posetleq\) is a partial order on \(\eventopts\)]
The reflexivity of \(\posetleq\) follows from \labelcref{axiom:event:idempotent}.
For its antisymmetry, let's assume that \(\eventopt[1]\posetleq\eventopt[2]\) and \(\eventopt[2]\posetleq\eventopt[1]\), so we infer from \cref{eq:event:ordering} that
\begin{equation*}
\group{\forall\opt\in\opts}\left\{
\begin{aligned}
\calledoff[1]{\opt}
&=\calledoff[2]{(\calledoff[1]{\opt})}\\
\calledoff[2]{\opt}
&=\calledoff[2]{(\calledoff[1]{\opt})}.
\end{aligned}
\right.
\end{equation*}
If we now let \(\opt\instantiateas\uniteventopt\), then we infer from \labelcref{axiom:event:unit:event} that, indeed, \(\eventopt[1]=\calledoff[2]{\eventopt[1]}=\eventopt[2]\).
For transitivity, assume that \(\eventopt[1]\posetleq\eventopt[2]\) and \(\eventopt[2]\posetleq\eventopt[3]\), then we infer from \cref{eq:event:ordering} that, for any \(\opt\in\opts\),
\begin{align*}
\calledoff[1]{\opt}
=\calledoff[1]{(\calledoff[2]{\opt})}
&=\calledoff[1]{(\calledoff[2]{(\calledoff[3]{\opt})})}\\
&=\calledoff[1]{(\calledoff[3]{\opt})}
\end{align*}
and similarly
\begin{align*}
\calledoff[1]{\opt}
=\calledoff[2]{(\calledoff[1]{\opt})}
&=\calledoff[3]{(\calledoff[2]{(\calledoff[1]{\opt})})}\\
&=\calledoff[3]{(\calledoff[1]{\opt})}.\qedhere
\end{align*}
\end{proof}

\begin{proof}[Proof that the poset \(\structure{\eventopts,\posetleq}\) has bottom~\(\nulleventopt\) and top~\(\uniteventopt\)]
Consider any \(\eventopt\in\eventopts\) and any \(\opt\in\opts\).
It follows from \labelcref{axiom:event:unit:event} that \(\eventopt\ast\group{\uniteventopt\ast\opt}=\eventopt\ast\opt=\uniteventopt\ast\group{\eventopt\ast\opt}\), and this tells us that, indeed, \(\eventopt\posetleq\uniteventopt\).
It also follows from \labelcref{axiom:event:linear,axiom:event:null:event} that \(\eventopt\ast\group{\nulleventopt\ast\opt}=\eventopt\ast0=0=\nulleventopt\ast\opt=\nulleventopt\ast\group{\eventopt\ast\opt}\), which tells us that, indeed, \(\nulleventopt\posetleq\eventopt\).
\end{proof}

\begin{proof}[Proof of \cref{prop:ordering:equivalent:statements}]
We give a circular proof that \labelcref{prop:event:ordering}\(\then\)\labelcref{prop:event:kernels}\(\then\)\labelcref{prop:event:nonpositivity}\(\then\)\labelcref{prop:event:ordering}.
For \labelcref{prop:event:ordering}\(\then\)\labelcref{prop:event:kernels}, assume that \(\eventopt[1]\posetleq\eventopt[2]\) and consider any \(\opt\in\opts\) such that \(\eventopt[2]\ast\opt=0\).
But then also, by assumption, \(\eventopt[1]\ast\opt=\eventopt[1]\ast\group{\eventopt[2]\ast\opt}=\eventopt[1]\ast0=0\), where the last equality follows from \labelcref{axiom:event:linear}.
That \labelcref{prop:event:kernels}\(\then\)\labelcref{prop:event:nonpositivity} is trivial, and that \labelcref{prop:event:nonpositivity}\(\then\)\labelcref{prop:event:ordering} is the essence of \labelcref{axiom:event:ordering}.
\end{proof}

\begin{proof}[Proof of \cref{eq:event:complement:properties}]
That \(\eventopt+\neg\eventopt=\uniteventopt\) follows from \labelcref{axiom:event:complement}\labelcref{it:event:complement:sum} and \labelcref{axiom:event:unit:event}, combined with the statement in \cref{eq:event:bijection}.
This implies that \(\neg\eventopt=\uniteventopt-\eventopt\), and therefore also \(\neg\group{\neg\eventopt}=\uniteventopt-\neg\eventopt=\uniteventopt-\group{\uniteventopt-\eventopt}=\eventopt\).
That \(\group{\neg\eventopt}\ast\eventopt=0\) follows from \labelcref{axiom:event:complement}\labelcref{it:event:complement:nullify} with \(\opt\instantiateas\uniteventopt\) and \labelcref{axiom:event:unit:event}.
If we now replace the event \(\eventopt\) by the event \(\neg\eventopt\) in the previous statement, we find that \(\group{\neg\group{\neg\eventopt}}\ast\group{\neg\eventopt}=0\), and therefore that finally also \(\eventopt\ast\group{\neg\eventopt}=0\).
\end{proof}

\begin{proof}[Proof of \cref{eq:background:compatibility:event}]
We have to prove that
\begin{equation*}
\posopts\cap\eventindifset=\emptyset
\ifandonlyif
\group{\forall\opt\in\opts}\group{\opt\optgt0\then\calledoff{\opt}\optgt0}.
\end{equation*}
Since the converse implication is trivial, we focus on the direct implication.
Assume that \(\posopts\cap\eventindifset=\emptyset\), and consider any \(\opt\in\posopts\), then we have to prove that \(\calledoff{\opt}\optgt0\).
It follows from \(\opt\optgt0\) that \(\eventopt\ast\opt\optgeq0\), by \labelcref{axiom:event:monotone}.
Since we know from the assumption that \(\opt\notin\eventindifset\) and therefore \(\eventopt\ast\opt\neq0\), this implies that, indeed, \(\eventopt\ast\opt\optgt0\).
\end{proof}

\begin{proof}[Proof that \(\uniteventopt\) is the only regular event]
It's clear from \labelcref{axiom:event:unit:event} that \(\uniteventopt\) satisfies the condition in \cref{eq:background:compatibility:event}, and therefore is a regular event.
Now consider any regular event~\(\eventopt\) and observe that \cref{eq:background:compatibility:event} and \labelcref{axiom:event:unit:event} imply that \(\group{\forall\opt\in\opts}\group{\uniteventopt\ast\opt\optgt0\then\calledoff{\opt}\neq0}\), or equivalently, \(\group{\forall\opt\in\opts}\group{\calledoff{\opt}=0\then\uniteventopt\ast\opt\not\optgt0}\).
This implies that \(\uniteventopt\posetleq\eventopt\), and therefore that \(\uniteventopt=\eventopt\), because we've proved above that \(\uniteventopt\) is the top of the poset~\(\structure{\eventopts,\posetleq}\).
\end{proof}

\begin{proof}[Proof that \(\nulleventopt=0\) is the only improper event]
First of all, \labelcref{axiom:event:null:event} implies that \(\nulleventopt\ast\opts=\set{0}\), which guarantees in turn that \(\posopts\cap\group{\nulleventopt\ast\opts}=\emptyset\), so the null event \(\nulleventopt\) isn't proper.
Next, consider any improper event~\(\eventopt\), meaning that \(\posopts\cap\eventopt\ast\opts=\emptyset\) and therefore \(\group{\forall\opt\optgeq0}\,\eventopt\ast\opt=0\), by \labelcref{axiom:event:monotone}.
But \labelcref{axiom:event:unit:event} then tells us that \(\uniteventopt\optgeq0\) and therefore also \(\eventopt=\eventopt\ast\uniteventopt=0\).
\end{proof}

\begin{proof}[Proof of the statement involving \cref{eq:desirability:indifference:revised}]
We start by proving that \(\desirset\altcondon\eventopt+\eventindifset=\desirset\altcondon\eventopt\).
It's clearly enough to show that \(\desirset\altcondon\eventopt+\eventindifset\subseteq\desirset\altcondon\eventopt\), so consider \(\
\opt\in\desirset\altcondon\eventopt\) and \(\altopt\in\eventindifset\), then \(\eventopt\ast\group{\opt+\altopt}=\eventopt\ast\opt+\eventopt\ast\altopt=\eventopt\ast\opt\in\desirset\), and therefore, indeed, \(\opt+\altopt\in\desirset\altcondon\eventopt\).
Since this implies that \(\posiof{\desirset\altcondon\eventopt\cup\eventindifset}=\desirset\altcondon\eventopt\cup\eventindifset\), we see that the assessment \(\Adelim{\desirset\altcondon\eventopt\cup\eventindifset}{-\desirset\altcondon\eventopt}\) is consistent because \(\desirset\altcondon\eventopt\cap\eventindifset=\emptyset\).
\cref{eq:model:closure} and some basic algebraic manipulations now tell us that, indeed, \(\Mclsbgof{\zeroM}{\Adelim{\desirset\altcondon\eventopt\cup\eventindifset}{-\desirset\altcondon\eventopt}}=\Adelim{\desirset\altcondon\eventopt\cup\eventindifset}{-\desirset\altcondon\eventopt}\).
\end{proof}

\begin{proof}[Proof of \Cref{prop:revision}]
It's clear that \labelcref{axiom:revision:1} is satisfied, as \(\reviseof{\M\given\eventM}\in\difMsabove{\eventbgM}\subseteq\closedMsabove{\zeroM}\).
Direct inspection based on \cref{eq:model:for:event,eq:revision:operator} also confirms that \labelcref{axiom:revision:2} holds.
For \labelcref{axiom:revision:3}, we first check the case that \(\eventopt=\uniteventopt\).
We recall from \cref{sec::events} that \(\desirset\altcondon\uniteventopt=\desirset\), \(\indifset[\uniteventopt]=\set{0}\) and \(\eventM[\uniteventopt]=\Adelim{\set{0}}{\emptyset}\subseteq\M\), so \(\reviseof{\M\given\eventM}=\Adelim{\desirset\cup\set{0}}{-\desirset}=\M=\expandof{\M\given\eventM}\).
For all other events, trivially, \(\reviseof{\M\given\eventM}\subseteq\Adelim{\opts}{\opts}=\expandof{\M\given\eventM}\).
For \labelcref{axiom:revision:4}, recall from the discussion  in \cref{sec::conditioning} [belief expansion] that \(\M\) and \(\eventM\) are only consistent when \(\eventopt=\uniteventopt\), and we've checked (above) that then \(\reviseof{\M\given\eventM}=\expandof{\M\given\eventM}\).
Axiom \labelcref{axiom:revision:5} is vacuously fulfilled because no model of the form \(\eventM=\Adelim{\eventindifset}{\emptyset}\) is inconsistent.
Since \(\eventM=\Mclsbgof{\zeroM}{\eventM}\), \labelcref{axiom:revision:6} holds too.
We'll tackle \labelcref{axiom:revision:7,axiom:revision:8} together.
The only case of interest is when \(\reviseof{\M\given\indexedeventM[1]}\) and \(\indexedeventM[2]\) are consistent, or equivalently, after some algebraic manipulations, when \(\desirset\altcondon{\eventopt[1]}\cap\indifset[2]=\emptyset\).
Because if not, then we'd have that \(\expandof{\reviseof{\M\given\indexedeventM[1]}\given\indexedeventM[2]}=\Adelim{\opts}{\opts}\), so \labelcref{axiom:revision:7} holds trivially; and \labelcref{axiom:revision:8} is satisfied vacuously.
So, if \(\desirset\altcondon{\eventopt[1]}\cap\eventindifset[2]=\emptyset\), we infer from \(\posopts\subseteq\desirset\) that in particular \(\group{\forall\opt\in\opts}\group{\calledoff[2]{\opts}=0 \then \calledoff[1]{\opt}\not\optgt0}\).
From \labelcref{axiom:event:ordering} we then infer that \(\eventopt[1]\posetleq\eventopt[2]\).
Using the first two equivalences in \cref{prop:ordering:equivalent:statements}, we see that \(\eventindifset[2]\subseteq\eventindifset[1]\), and therefore \(\indexedeventM[2]\subseteq\indexedeventM[1]\).
This implies that, on the one hand, \(\indexedeventM[1]\cup\indexedeventM[2]=\indexedeventM[1]\) and \(\reviseof{\M\given\indexedeventM[1]\cup\indexedeventM[2]}=\reviseof{\M\given\indexedeventM[1]}\).
On the other hand, we find that \(\indexedeventM[2]\subseteq\indexedeventM[1]\subseteq\reviseof{\M\given\indexedeventM[1]}\) and therefore \(\expandof{\reviseof{\M\given\indexedeventM[1]}\given\indexedeventM[2]}=\reviseof{\M\given\indexedeventM[1]}\).
This proves both \labelcref{axiom:revision:7} and \labelcref{axiom:revision:8}.
\end{proof}

\begin{proof}[Proof of \Cref{prop:contraction}]
We start by observing that both \(\desirset\altcondon\neg\eventopt\cap\desirset\) and \(\desirset\cap\indifset[\neg\eventopt]\) are convex cones, and that
\begin{multline*}
\contractof{\M\given\eventM}\\
=\Adelim{(\desirset\altcondon\neg\eventopt\cap\desirset)\cup(\desirset\cap\indifset[\neg\eventopt])\cup\{0\}}{-(\desirset\altcondon\neg\eventopt\cap\desirset)}.
\end{multline*}
\labelcref{axiom:contraction:1} holds because \(\contractof{\M\given\eventM}\) is defined in \cref{eq:contraction:operator} as a meet of two closed models \(\M\) and \(\reviseof{\M\given\neg\eventM}\) that respect \(\zeroM\) [\(\reviseof{\M\given\neg\eventM}\in\closedMsabove{\zeroM}\) by \Cref{prop:revision}].
\labelcref{axiom:contraction:2} holds because, clearly, \(\desirset\cap\desirset\altcondon\neg\eventopt\subseteq\desirset\) and \(\desirset\cap\indifset[\neg\eventopt]\subseteq D\).
We saw in \cref{sec::conditioning} that \(\neg\eventM\) and \(\M\) are only consistent when \(\neg \eventM\) corresponds to the regular event~\(\uniteventopt\), so when \(\eventopt=\nulleventopt=0\).
Since then \(\desirset\altcondon\neg\eventopt=\desirset\altcondon\uniteventopt=\desirset\) and \(\indifset[{\neg\eventopt}]=\indifset[\uniteventopt]=\set{0}\), we can come to the conclusion that \(\contractof{\M\given\eventM}=\contractof{\M\given\eventM[\nulleventopt]}=\Adelim{\desirset\cup\set{0}}{-\desirset}=\M\), so \labelcref{axiom:contraction:3} holds.
The only event~\(\eventopt\) for which \(\eventM\subseteq\contractof{\M\given\eventM}\), and therefore \(\eventM\subseteq\M\), is the regular unit event \(\uniteventopt\), which is outside the scope of the proposition, so \labelcref{axiom:contraction:4} holds vacuously.
\labelcref{axiom:contraction:5} is also vacuously fulfilled, for the same reason.
Since \(\Mclsbgof{\zeroM}{\eventM}=\eventM\), \labelcref{axiom:contraction:6} is also obeyed.
\end{proof}

\begin{proof}[Proof of \Cref{prop:identity:levi}]
First, observe that the assessment \(\contractof{\M\given\neg\eventM}\union\eventM=\Adelim{\group{\desirset\altcondon\eventopt\cap\desirset}\cup\eventindifset}{-\group{\desirset\altcondon\eventopt\cap\desirset}}\) is consistent, because
\begin{align*}
\posiof{\group{\desirset\altcondon\eventopt\cap\desirset}\cup\eventindifset}
&=\group{\group{\desirset\altcondon\eventopt\cap\desirset}+\eventindifset}\cup\eventindifset
\end{align*}
and \(\desirset\altcondon\eventopt\cap\eventindifset=\emptyset\).
This tells us that, using \cref{eq:model:closure}, and after some algebraic manipulations,
\begin{multline*}
\expandof{\contractof{\M\given\neg\eventM}\given\eventM}\\
\begin{aligned}
&=\Mclsbgof{\zeroM}{\contractof{\M\given\neg\eventM}\union\eventM}\\
&=\Adelim{\group{\group{\desirset\altcondon\eventopt\cap\desirset}+\eventindifset}\cup\eventindifset}{-\group{\desirset\altcondon\eventopt\cap\desirset+\eventindifset}},
\end{aligned}
\end{multline*}
so it's enough to prove that \(\group{\desirset\altcondon\eventopt\cap\desirset}+\eventindifset=\desirset\altcondon\eventopt\).
On the one hand, \(\group{\desirset\altcondon\eventopt\cap\desirset}+\eventindifset\subseteq\desirset\altcondon\eventopt+\eventindifset=\desirset\altcondon\eventopt\).
For the converse inclusion, consider that for all \(\opt\in\desirset\altcondon\eventopt\), \(\opt=\calledoff{\opt}+\group{\opt- \calledoff{\opt}}\), where \(\opt-\calledoff{\opt}\in\eventindifset\) and \(\calledoff{\opt}\in\desirset\).
\end{proof}

\begin{proof}[Proof that \labelcref{axiom:event:ordering} holds in a quantum context]
Assume that \(\group{\forall\measurement{A}\in\measurements}\group{\projector[2]\measurement{A}\projector[2]=\zero\then\projector[1]\measurement{A}\projector[1]\not\weakgt\zero}\), then we must prove that \(\projector[1]=\projector[2]\projector[1]=\projector[1]\projector[2]\).
Consider any \(\fket\in\statespace\) and let \(\subspace\coloneqq\linspanof{\set{\fket}}\), the linear space spanned by \(\fket\).
Observe that for any \(k\in\set{1,2}\),
\begin{equation}\label{eq:ordering:projection}
\projector[k]\projector\projector[k]\weakgeq\zero
\text{ and }
\group{\projector[k]\projector\projector[k]=\zero\ifandonlyif\fket\in\subspace[k]^\perp}.
\end{equation}
Now assume that \(\fket\in\subspace[2]^\perp\), then we infer from the assumption and \Cref{eq:ordering:projection} that \(\fket\in\subspace[1]^\perp\), implying that \(\subspace[2]^\perp\subseteq\subspace[1]^\perp\), or in other words,
\[
\projectoron{\subspace[2]^\perp}\projectoron{\subspace[1]^\perp}
=\projectoron{\subspace[1]^\perp}\projectoron{\subspace[2]^\perp}
=\projectoron{\subspace[2]^\perp},
\text{ with }\projectoron{\subspace[k]^\perp}=\identity-\projectoron{\subspace[k]}.
\]
A little algebra now leads to the desired equalities.
\end{proof}

\ifarXiv
\begin{proof}[Proof that \labelcref{axiom:event:kernel:sum} holds in a classical probability context]
It's clearly enough to prove that \(\indifset[{\event[1]}]+\indifset[{\event[2]}]=\indifset[{\event[1]\cap\event[2]}]\) for all \(\event[1],\event[2]\subseteq\states\).
First, assume that \(\gbl\in\indifset[{\event[1]}]+\indifset[{\event[2]}]\), so there are \(\gbl_1\in\indifset[{\event[1]}]\) and \(\gbl_2\in\indifset[{\event[2]}]\) such that \(\gbl=\gbl_1+\gbl_2\).
But then we know that \(\indevent[1]\gbl_1=\indevent[2]\gbl_2=0\) and therefore also
\begin{equation*}
\indof{\event[1]\cap\event[2]}\gbl
=\indevent[1]\indevent[2]\group{\gbl_1+\gbl_2}
=\indevent[2]\indevent[1]\gbl_1+\indevent[1]\indevent[2]\gbl_2
=0,
\end{equation*}
so \(f\in\indifset[{\event[1]\cap\event[2]}]\), and therefore \(\indifset[{\event[1]}]+\indifset[{\event[2]}]\subseteq\indifset[{\event[1]\cap\event[2]}]\).
For the converse inclusion, assume that \(f\in\indifset[{\event[1]\cap\event[2]}]\), and let \(\gbl_1\coloneqq\nicefrac12\group{1-\indevent[1]+\indevent[2]}\gbl\) and \(\gbl_2\coloneqq\nicefrac12\group{1-\indevent[2]+\indevent[1]}\gbl\).
Then \(\indevent[1]\gbl_1=\indevent[2]\gbl_2=0\) and \(f=\gbl_1+\gbl_2\).
\end{proof}
\fi

\ifarXiv
\begin{proof}[Proof that \labelcref{axiom:event:kernel:sum} holds in a quantum probability context]
Let \(\projectoron{k}\coloneqq\projector[k]\) and \(\projectoron{1,2}\coloneqq\projectoron{\subspace[1]\cap\subspace[2]}\), then \(\smash{\indifset[\projectoron{1}]}+\smash{\indifset[\projectoron{2}]}\subseteq\smash{\indifset[\projectoron{1,2}]}\) follows trivially from \(\projectoron{k}=\projectoron{k}\projectoron{1,2}=\projectoron{1,2}\projectoron{k}\).
So it now remains to prove that \(\smash{\indifset[\projectoron{1,2}]}\subseteq\smash{\indifset[\projectoron{1}]}+\smash{\indifset[\projectoron{2}]}\).
Consider any \(\measurement{A}\) such that \(\projectoron{1,2}\measurement{A}\projectoron{1,2}=\zero\), then we must find \(\measurement{B}\) and \(\measurement{C}\) such that \(\projectoron{1}\measurement{B}\projectoron{1}=\zero\), \(\projectoron{2}\measurement{C}\projectoron{2}=\zero\) and \(\measurement{A}=\measurement{B}+\measurement{C}\).

By von Neumann's alternating orthogonal projection result \cite[Theorem 13.7]{neumann1950}, \(\projectoron{1,2}=\lim_{n\to\infty}\group{\projectoron{1}\projectoron{2}}^n=\lim_{n\to\infty}\group{\projectoron{2}\projectoron{1}}^n\).
Let
\begin{equation*}
\measurement{B}_k
\coloneqq\group{\projectoron{2}\projectoron{1}}^k\projectoron{2}\measurement{A}\projectoron{2}\group{\projectoron{1}\projectoron{2}}^k
-\group{\projectoron{1}\projectoron{2}}^{k+1}\measurement{A}\group{\projectoron{2}\projectoron{1}}^{k+1},
\end{equation*}
then \(\projectoron{1}\measurement{B}_k\projectoron{1}=\zero\).
Since \(\measurement{B}'_n\coloneqq\sum_{k=0}^n\measurement{B}_k\) converges to some \(\measurement{B}\) in \(\measurements\) [essentially because \(\projectoron{1}\projectoron{2}\) and \(\projectoron{2}\projectoron{1}\) can be written as orthogonal sums of an identity on \(\subspace[1]\cap\subspace[2]\) and a strict contraction on its orthogonal complement; see \cite[Thm.~3.1]{kuo1989:factorisation}], we find that \(\projectoron{1}\measurement{B}\projectoron{1}=\zero\).
We are done if we can prove that \(\projectoron{2}\group{\measurement{A}-\measurement{B}}\projectoron{2}=\zero\).
Check that
\begin{multline*}
\projectoron{2}\group{\measurement{A}-\measurement{B}'_n}\projectoron{2}
=\group{\projectoron{2}\projectoron{1}}^{n+1}\projectoron{2}\measurement{A}\projectoron{2}\group{\projectoron{1}\projectoron{2}}^{n+1}\\
\to\projectoron{1,2}\projectoron{2}\measurement{A}\projectoron{2}\projectoron{1,2},
\end{multline*}
and therefore, indeed,
\begin{equation*}
\projectoron{2}\group{\measurement{A}-\measurement{B}}\projectoron{2}
=\projectoron{1,2}\projectoron{2}\measurement{A}\projectoron{2}\projectoron{1,2}
=\projectoron{2}\projectoron{1,2}\measurement{A}\projectoron{1,2}\projectoron{2}
=\zero.\qedhere
\end{equation*}
\end{proof}
\fi

\additionalinfo

\begin{acknowledgements}
We'd like to express our debt and gratitude to Erik Quaeghebeur and Alessandro Facchini for inspiring discussions over the past few years, and to Jasper De Bock for hearing us out when we needed him to, as well as for his always accurate and also always deeply appreciated criticism.
Arthur Van Camp's enthusiasm helped us squash a few bugs in the manuscript.
Funding for Gert de Cooman's research is partly covered by Ghent University's pioneering non-competitive research funding initiative.

We also want to express our appreciation to four anonymous reviewers, who amongst them made several suggestions for improvement, and pointed out further connections to the existing literature.
\end{acknowledgements}

\begin{authorcontributions}
This paper was born from long and intensive discussions between all three authors, and all three of us have made essential contributions to its ideas and content.
As far as its execution is concerned, Kathelijne and Gert did most of the writing, while Keano focused on reviewing.
The ideas and material in the section on events were initially based on a chapter in a book on imprecise probabilities that Gert is currently writing, but our cooperation here has also led to new ideas and insights there as well.
\end{authorcontributions}

\printbibliography

@article{alchourron1985,
  author  = {Alchourrón, Carlos  and Gärdenfors, Peter and Makinson, David},
  doi     = {10.2307/2274239},
  journal = {Journal of Symbolic Logic},
  pages   = {510--530},
  title   = {On the logic of theory change: Partial meet contraction and revision functions},
  volume  = 50,
  year    = {1985},
}

@article{benavoli2016:quantum_2016,
  archiveprefix = {arXiv},
  author        = {Benavoli, Alessio and Facchini, Alessandro and Zaffalon, Marco},
  doi           = {10.1103/PhysRevA.94.042106},
  eprint        = {1605.08177},
  journal       = {Physical Review A},
  month         = oct,
  number        = {4},
  primaryclass  = {quant-ph},
  shorttitle    = {Quantum mechanics},
  title         = {Quantum mechanics: the Bayesian theory generalised to the space of {Hermitian} matrices},
  volume        = {94},
  year          = {2016},
}

@article{cooman2003a,
  author  = {\VAN{Cooman}{De}{de} Cooman, Gert},
  doi     = {10.1007/s10472-005-9006-x},
  journal = {Annals of Mathematics and Artificial Intelligence},
  pages   = {5--34},
  title   = {Belief models: an order-theoretic investigation},
  volume  = {45},
  year    = {2005},
}

@article{cooman2010,
  author  = {\VAN{Cooman}{De}{de} Cooman, Gert and Quaeghebeur, Erik},
  doi     = {10.1016/j.ijar.2010.12.002},
  journal = {International Journal of Approximate Reasoning},
  note    = {Special issue in honour of Henry E.~Kyburg, Jr.},
  number  = {3},
  pages   = {363--395},
  title   = {Exchangeability and sets of desirable gambles},
  volume  = {53},
  year    = {2012},
}

@article{debock2016:partial:exchangeability,
  author   = {De Bock, Jasper and Van Camp, Arthur and Diniz, Márcio A. and \VAN{Cooman}{De}{de} Cooman, Gert},
  doi      = {10.1016/j.fss.2014.10.027},
  issn     = {0165-0114},
  journal  = {Fuzzy Sets and Systems},
  language = {English},
  pages    = {1--30},
  title    = {Representation theorems for partially exchangeable random variables},
  url      = {http://www.sciencedirect.com/science/article/pii/S0165011414004837},
  volume   = {284},
  year     = {2016},
}

@article{decooman2015:coherent:predictive:inference,
  author  = {\VAN{Cooman}{De}{de} Cooman, Gert and De Bock, Jasper and Diniz, M\'arcio Alves},
  doi     = {10.1613/jair.4490},
  journal = {Artificial Intelligence Journal},
  pages   = {1--95},
  title   = {Coherent predictive inference under exchangeability with imprecise probabilities},
  volume  = {52},
  year    = {2015},
}

@article{devos2023:indistinguishability,
  author  = {De Vos, Keano and \VAN{Cooman}{De}{de} Cooman, Gert and De Bock, Jasper},
  journal = {Proceedings of Machine Learning Research},
  pages   = {177--188},
  title   = {Indistinguishability through Exchangeability in Quantum Mechanics?},
  volume  = {215},
  year    = {2023},
}

@article{devos2025:isipta,
  author = {De Vos, Keano and \VAN{Cooman}{De}{de} Cooman, Gert},
  note   = {Submitted to ISIPTA 2025},
  title  = {Conditioning through indifference in quantum mechanics},
  year   = {2025},
}

@article{finetti1937,
  author  = {{d}e Finetti, Bruno},
  journal = {Annales de l'Institut Henri Poincar\'e},
  note    = {{E}nglish translation in \cite{kyburg1964}},
  pages   = {1--68},
  title   = {La pr\'evision: ses lois logiques, ses sources subjectives},
  volume  = {7},
  year    = {1937},
}

@book{finetti19745,
  address   = {Chichester},
  author    = {{d}e Finetti, Bruno},
  note      = {{E}nglish translation of \cite{finetti1970}, two volumes},
  publisher = {John Wiley \& Sons},
  title     = {Theory of Probability: A Critical Introductory Treatment},
  year      = {1974--1975},
}

@book{gardenfors1988,
  address   = {Cambridge, MA},
  author    = {G\"ardenfors, Peter},
  publisher = {The MIT Press},
  title     = {Knowledge in Flux -- Modeling the Dynamics of Epistemic States},
  year      = {1988},
}

@article{kuo1989:factorisation,
  author  = {Kuo, Kung Hwang and Wu, Pei Yuan},
  doi     = {10.1090/s0002-9939-1989-0977922-1},
  journal = {Proceedings of the American Mathematical Society},
  number  = {2},
  pages   = {263--272},
  title   = {Factorization of matrices into partial isometries},
  volume  = {105},
  year    = {1989},
}

@book{neumann1950,
  author    = {von Neumann, John},
  publisher = {Princeton University Press},
  title     = {Functional Operators, Vol II: The Geometry of Orthogonal Spaces},
  year      = {1950},
}

@article{quaeghebeur2015:statement,
  author  = {Quaeghebeur, Erik and \VAN{Cooman}{De}{de} Cooman, Gert and Hermans, Filip},
  doi     = {10.1016/j.ijar.2014.12.003},
  journal = {International Journal of Approximate Reasoning},
  pages   = {69--102},
  title   = {Accept {\&} reject statement-based uncertainty models},
  volume  = {57},
  year    = {2015},
}

@book{rott2001,
  author    = {Rott, Hans},
  publisher = {Oxford University Press},
  title     = {Change, Choice and Inference: A Study of Belief Revision and Nonmonotonic Reasoning},
  year      = {2001},
}

@book{troffaes2013:lp,
  author    = {Troffaes, Matthias C. M. and \VAN{Cooman}{De}{de} Cooman, Gert},
  publisher = {Wiley},
  title     = {Lower Previsions},
  year      = 2014,
}

@book{walley1991,
  address   = {London},
  author    = {Walley, Peter},
  publisher = {Chapman and Hall},
  title     = {Statistical Reasoning with Imprecise Probabilities},
  year      = {1991},
}

@article{walley2000,
  author  = {Walley, Peter},
  doi     = {10.1016/S0888-613X(00)00031-1},
  journal = {International Journal of Approximate Reasoning},
  pages   = {125--148},
  title   = {Towards a unified theory of imprecise probability},
  volume  = {24},
  year    = {2000},
}

@techreport{williams1975,
  address     = {University of Sussex, UK},
  author      = {Williams, Peter M.},
  institution = {School of Mathematical and Physical Science},
  title       = {Notes on conditional previsions},
  year        = {1975},
}

@article{williams2007,
  author  = {Williams, Peter M.},
  doi     = {10.1016/j.ijar.2006.07.019},
  journal = {International Journal of Approximate Reasoning},
  note    = {Revised journal version of \cite{williams1975}},
  pages   = {366--383},
  title   = {Notes on conditional previsions},
  volume  = {44},
  year    = {2007},
}

\end{document}